\documentclass[journal,twocolumn,10pt]{IEEEtran}
\usepackage{amsfonts}
\usepackage{amsmath,cases,amssymb}
\usepackage[dvips]{graphicx}
\usepackage{subfigure}
\usepackage{setspace}
\usepackage{fixltx2e}
\usepackage{cite,color}
\usepackage{epsfig}
\usepackage{url}
\usepackage{algorithm}
\usepackage{algorithmic}
\usepackage{multirow}
\usepackage{mathtools}
\usepackage{lipsum}
\usepackage{amsmath}
\usepackage{dblfloatfix}
\usepackage{epstopdf}

\newcommand{\ds}{\displaystyle}

\newcommand{\bz}{\mathbf{z}}
\newcommand{\br}{\mathbf{r}}

\newcommand{\clC}{{\mathcal C}}
\newcommand{\clL}{{\mathcal L}}
\newcommand{\clN}{{\mathcal N}}

\newcommand{\clK}{{\mathcal K}}

\newcommand{\clD}{{\mathcal D}}

\newcommand{\bw}{\mathbf{w}}
\newcommand{\bm}{\mathbf{m}}

\ifCLASSINFOpdf

\else
\fi

\begin{document}
\title{Distributed Semi-supervised Fuzzy Regression with Interpolation Consistency Regularization}

\author{Ye Shi, \textit{Member, IEEE}, Leijie Zhang, Zehong Cao, \textit{Member, IEEE}, and M. Tanveer, \textit{Senior Member, IEEE}, Chin-Teng Lin \textit{Fellow, IEEE}
\thanks{Ye Shi and Leijie Zhang contributed equally to this work. This work was supported in part by the Australian Research Council (ARC) under Discovery Projects DP180100670 and DP180100656 and in part by the Office of Naval Research Global, US, under Cooperative Agreement Number ONRG-NICOP-N62909-19-1-2058. We would also like to thank the NSW Defence Innovation Network and NSW State Government of Australia for financial support of this project through grant DINPP2019 S1-03/09. (Corresponding author: Chin-Teng Lin)}
\thanks{Ye Shi, Leijie Zhang and Chin-Teng Lin are with the School of Computer Science, University of Technology, Sydney, NSW 2007, Australia. Email:
ye.shi-1@uts.edu.au, Leijie.Zhang@student.uts.edu.au, Chin-Teng.Lin@uts.edu.au.}
\thanks{Zehong Cao is with the School of Computer Science, University of Technology Sydney, Sydney, NSW 2007, Australia and the Department of Discipline of ICT, University of Tasmania, Hobart, TAS 7001, Australia. Email: zhcaonctu@gmail.com.}
\thanks{M. Tanveer is with the Discipline of Mathematics, Indian Institute of Technology Indore, Simrol, Indore, 453552, India. Email: mtanveer@iiti.ac.in.}
}

\maketitle

\begin{abstract}
Recently, distributed semi-supervised learning (DSSL) algorithms have shown their effectiveness in leveraging unlabeled samples over interconnected networks, where agents cannot share their original data with each other and can only communicate non-sensitive information with their neighbors. However, existing DSSL algorithms cannot cope with data uncertainties and may suffer from high computation and communication overhead problems. To handle these issues, we propose a distributed semi-supervised fuzzy regression (DSFR) model with fuzzy if-then rules and interpolation consistency regularization (ICR). The ICR, which was proposed recently for semi-supervised problem, can force decision boundaries to pass through sparse data areas, thus increasing model robustness. However, its application in distributed scenarios has not been considered yet. In this work, we proposed a distributed Fuzzy C-means (DFCM) method and a distributed interpolation consistency regularization (DICR) built on the well-known alternating direction method of multipliers to respectively locate parameters in antecedent and consequent components of DSFR. Notably, the DSFR model converges very fast since it does not involve back-propagation procedure and is scalable to large-scale datasets benefiting from the utilization of DFCM and DICR. Experiments results on both artificial and real-world datasets show that the proposed DSFR model can achieve much better performance than the state-of-the-art DSSL algorithm in terms of both loss value and computational cost. Our code is available online\footnote{\url{https://github.com/leijiezhang/DSFR}}.
\end{abstract}


\section{Introduction}
Over the past few decades, semi-supervised learning (SSL) algorithms have been thoroughly investigated  for ways to use both labeled and unlabeled data to train better models \cite{zhu2005semi, yarowsky1995unsupervised, riloff2003learning,rosenberg2005semi, blum1998combining, anis2018sampling, xu2019enhanced, jin2018regularized}. The different categories of SSL algorithms generally include self-training \cite{yarowsky1995unsupervised, riloff2003learning,rosenberg2005semi}, co-training \cite{blum1998combining}, graph-based methods \cite{anis2018sampling, xu2019enhanced, jin2018regularized} and MixUp-based methods \cite{zhang2017mixup,verma2019interpolation}, etc. Most of these methods work by adding regularization terms that deal with unlabeled samples in the loss functions. Further, most of them only work in centralized scenarios, where training data is processed at a central node. However, rising concerns over data privacy and security  \cite{predd2006distributed, barni2011privacy, pathak2012privacy, pathak2011privacy} has made distributed computing a far more popular paradigm. Consequently, in many real-world scenarios \cite{mohassel2017secureml, tuttle2018facebook, yan2012distributed}, the available training data is spread across interconnected networks comprising multiple agents. Typically, these agents are not allowed to share the data they have with others and can only communicate non-sensitive information to their neighbors. As such, the training data must be stored and processed on multiple local nodes instead of in one central place. To date, only a few researchers have investigated distributed semi-supervised learning (DSSL) \cite{chang2017distributed,xie2019distributed,xie2019distributedelm}, and none have considered data uncertainty in the training samples. Both properties are common in real-world datasets, so a framework that considers these factors would invariably increase model performance, especially with complex data.

Data uncertainty inherently exists during the process of data collection due to measurement errors, incomplete knowledge and subject difference. Generally, there are two types of data uncertainty: epistemic uncertainty and aleatoric uncertainty. The former usually comes from observations or linguistic ambiguity in knowledge while the latter results mainly from to inherent variability of physical systems. One of the best methods for dealing with incomplete or uncertain information is fuzzy inference systems \cite{jang1993anfis, el2018fuzzy,deng2015transfer} that rely on fuzzy logic.
As learning machines that find the parameters of fuzzy systems (i.e., fuzzy sets, fuzzy rules), fuzzy neural networks (FNN) \cite{couso2019fuzzy, fu2011fuzzy, chen2019prediction}
comprise an antecedent and a consequent component that offers a specific architecture for tackling data uncertainty.  Actually, fuzzy systems have been introduced to SSL methods for years and has found their way onto many directions \cite{yan2013fuzzy, poria2012fuzzy, zhou2014fuzzy, zhang2020robust, jiang2017seizure}. Fuzzy SSL algorithms, however, are still in their infancy. They usually rely heavily on human knowledge and can only be processed in a centralized way.

In search of a fuzzy SSL method that can better utilize unlabeled samples, we were motivated by a recently developed technique called interpolation consistency regularization (ICR) \cite{verma2019interpolation}. Unlike other SSL methods, which essentially use unlabeled data to supplement the available training data, ICR expands the sample space to capture more extrinsic information. ICR not only helps to train a better estimator based on augmented training samples, but also encourages consistency between predictions based on the augmented samples, i.e., $f(\lambda x_i + (1-\lambda) x_j)$, and interpolated predictions based on those samples, i.e., $\lambda f(x_i) + (1-\lambda)f(x_j)$. Further, ICR can push the decision boundaries toward low-density areas, which increases model robustness and generalization performance. Currently, ICR methods have been widely used in semi-supervised classification tasks with backpropagation training schemes \cite{zhang2017mixup,verma2019interpolation,berthelot2019mixmatch}, but seldom with semi-supervised regression tasks.

In this paper, we designed a distributed semi-supervised fuzzy regression (DSFR) model with fuzzy if-then rules and ICR to handle data uncertainty and reduce computation and communication overheads in interconnected networks. A distributed Fuzzy C-means (DFCM) algorithm locates the parameters in an antecedent component and a distributed interpolation consistency regularization (DICR) algorithm obtains the parameters in a consequent component. Both the DFCM algorithm and the DICR algorithm are implemented following the well-known alternating direction method of multipliers (ADMM) \cite{boyd2011convex} to guarantee consensus among all local agents. Data is only ever processed locally by the agent that owns the data, and very little information is passed between neighbors, all of which is non-sensitive. Notably, the DSFR model converges very quickly since it does not involve a backpropagation procedure. Benefiting from DFCM and DICR, it also scales well to large datasets.
Thus, the main contributions of this paper include:
\begin{itemize}
\item A novel DSFR model with ICR that handles data uncertainty and has lower computation and communication overheads in interconnected networks than existing DSSL algorithms.
\item A  DFCM algorithm to locate parameters in the antecedent component of the DSFR model. The DFCM can be used directly with both labeled and unlabeled training data available over interconnected networks.
\item A DICR algorithm to obtain parameters in the consequent component of the DSFR model. This is the first implementation to extend ICR to a distributed and semi-supervised scenario. In contrast to existing DDSL algorithms, such as graph-based DDSL \cite{xie2019distributed,xie2019distributedelm}, DICR results in smaller loss values and enjoys much greater scalability.

\end{itemize}
The remainder of the paper is organized as follows: Section II describes related work; Sections III and IV are devoted to centralized and distributed semi-supervised fuzzy regression with ICR, respectively; and Section V presents the experiments conducted to evaluate and verify the proposed model.

\section{Related work}

\subsection{Distributed Learning}
Distributed learning algorithms \cite{forero2010consensus,qin2016distributed,bi2015distributed,ye2019decentralized,scardapane2016decentralized} have been successfully applied to many real-world applications, including wireless sensor networks \cite{qin2016distributed} and privacy-preserving \cite{liu2015smc}. Qin et al. \cite{qin2016distributed} combined graph theory with the distributed consensus theory of multi-agent systems to design two decentralized algorithms for dealing with distributed problems in wireless sensor networks. One algorithm is based on K-means, the other on FCM. Targeting privacy preservation, Liu et al. \cite{liu2015smc} devised a novel shadow coding schema to save and recover privacy information over distributed networks. Two interesting works in \cite{ye2020distributed1} and \cite{ye2020distributed2} provided novel learning algorithms for distributed games. Deep learning solutions involving consensus models over distributed deep architectures have also been proposed with the aim of improving performance \cite{taylor2016training,leng2018extremely}. For example, Forero et al. \cite{forero2010consensus} proposed a fully distributed method based on support vector machine algorithms by applying an ADMM strategy to obtain optimal global parameters without the need to exchange and process information through a central communications unit. Ye et al. \cite{ye2019decentralized} later presented a decentralized ELM algorithm that combines the Jacobian and Gauss-Seidel Proximal ADMM methods. Obviously, all these distributed learning algorithms do well in alleviating privacy concerns and dealing with decentralized scenarios. However, none address uncertainty in the training data.

There are also several examples of fuzzy algorithms for distributed learning \cite{fierimonte2016distributed, fierimonte2017distributed, ye2020consensus, dang2020transfer}. For example, Fierimonte  et al. \cite{fierimonte2016distributed} developed a decentralized FNN with random weights, where parameters in the fuzzy membership functions are chosen randomly as opposed to being trained. In subsequent work, they introduced an online implementation of the same FNN structure \cite{fierimonte2017distributed}. Notably, a random method of identifying parameters can result in very large deviations in accuracy during the learning process. Additionally, the algorithms are only applied in the consequent layers of the FNN \cite{fierimonte2016distributed,fierimonte2017distributed}, which means that, strictly speaking, this decentralized FNN model is only partially distributed. A more recent proposition by Shi et al. \cite{ye2020consensus} involves a distributed FNN with a consensus learning strategy. A novel method of distributed clustering optimize the parameters in the antecedent layer, while a similar method of distributed parameter learning does the same for the consequent layer. This solution successfully manages data uncertainty in a distributed setting, but it does not consider unlabeled samples. Dang et al. \cite{dang2020transfer} proposed a transfer fuzzy clustering method to enable neighboring agents to learn from each other collaboratively. Similar to distributed learning, multi-view learning converges to an optimal estimator by collaboratively learning from multiple datasets. By applying a large margin learning mechanism, \cite{jiang2016realizing} proposed a two-view fuzzy model that collaboratively learning from each other. A multi-view fuzzy logic system is introduced in \cite{zhang2018multiview} by adopting nonnegative matrix factorization to build a hidden space in order to extract useful information shared among different datasets.

\subsection{Semi-supervised learning}
SSL algorithms \cite{yarowsky1995unsupervised, blum1998combining, riloff2003learning} were developed to improve model performance with additional training data in the form of unlabeled samples, while still leveraging the accuracy afforded by labeled samples. The first tools to solve SSL problems were self-training \cite{yarowsky1995unsupervised, riloff2003learning} methods, which  uses unlabeled samples in an iterative two-part procedure. First, the model is trained on a set of only labeled samples. The trained model is then used to predict the labels of unlabeled data. High-confidence predictions are then added to the training set and the model is retrained.
An alternative strategy is co-training \cite{blum1998combining}, which takes a similar iterative approach to self-training. The difference is that co-training algorithms maintain two separate estimators which work together on different training subsets and teach each other during the learning process. Subsequently, generative models \cite{zhu2005semi, nigam2000text, baluja2000using} were  proposed.  These algorithms assume that the model can represent hybrid distributions as identified in a set of unlabeled samples.

More recently, an increasing number of researchers are incorporating regularization terms into the loss function of SSL algorithms as a way to extract more useful information from the unlabeled data. Generally, these regularization terms fall into one of three categories. These are: traditional regularization \cite{zhang2018three},  which summarizes and transfers traditional SSL models into regularization terms; consistency regularization \cite{cirecsan2010deep}, which forces the predictions generated to be low-entropy so the decision boundary does not land in a dense sample area; and entropy minimization \cite{grandvalet2005semi, lee2013pseudo}, which guarantees that the distribution of the augmented dataset will be the same as the original.

To implement SSL in deep architecture, an interpolation consistency training procedure, Verma et al.  \cite{verma2019interpolation} proposed a solution inspired by MixUp \cite{zhang2017mixup} that learns a decision boundary which ensures consistency between predictions based on the interpolated samples and interpolations based on the resulting predictions. Meanwhile, Berthelot et al. \cite{berthelot2019mixmatch} was developing MixMatch – the current state-of-the-art in model performance. MixMatch basically combines many of the recent dominant SSL mechanisms, including entropy minimization and MixUp tools. However, all these algorithms are designed for centralized scenarios and cannot applied directly to decentralized problem solving.

Only a few DSSL methods have been proposed \cite{chang2017distributed, xie2019distributed, xie2019distributedelm} Chang et al. \cite{chang2017distributed} uses unlabeled data to reduce distribution errors and applied time-consuming kernel ridge regression on distributed nodes, after which the weighted average of those nodes’ outputs is calculated to obtain a final estimator. Taking advantage of a wavelet neural network, Xie et al. \cite{xie2019distributed} devised a new DSSL scheme that incorporates a graph-based regularization term into a distributed loss function. However, the process involves constructing a relationship graph of all the sample points, which is quite time-consuming, especially with large-scale datasets. In pursuit of better efficiency, the researchers later modified their information sharing strategy to use an event-triggered communication scheme \cite{xie2019distributedelm}, and in a later work still, they updated the objective function  to reduce the complexity of loss function by avoiding the twice continuously differential in their previous two models \cite{xie2020distributed}. While there are many ways reduce computing time, high computational overhead is an inherent problem with graph-based regularization that cannot be solved completely. Ultimately, the choice of DSSL strategy comes down to one of either less robustness or more time-consumption.

\section{Centralized semi-supervised fuzzy regression}
This section sets out the formulation for the centralized semi-supervised fuzzy regression (CSFR) model with ICR. Notably, CSFR is sequentially trained by an unsupervised structure learning and a semi-supervised parameter learning.

\subsection{Fuzzy inference system}
Let us briefly describe the fuzzy inference system based on a first-order Takagi-Sugeno (T-S) method. Consider the estimation of a scalar output $y\in \mathbb{R}$ from a $D$-dimensional input $x = [x_1,x_2,\cdots,x_D]$. The $k$-th fuzzy rule can be represented as
\begin{center}
  Rule $k$: IF $x_1$ is $A_{k1}$ and $\cdots$ and $x_D$ is $A_{kd}$\\
  Then $y$ = $w_{k0} + \sum_{j=1}^{D}w_{kj}x_j$
\end{center}
where $A_{kj}$ is a Gaussian fuzzy set with the following membership function:
\begin{equation}\label{Gaussian_A}
\varphi_{kj}(x_j) = \mbox{exp}\left[-\left(\frac{x_j-m_{kj}}{\sigma_{kj}}\right)^2\right]
\end{equation}
where $m_{kj}$ and $\sigma_{kj}$ are respectively the mean value and standard deviation.

The firing strength for each fuzzy rule is
\begin{equation}\label{firing_norm}
\bar{\phi}_k(x) = \frac{\prod_{j=1}^{D}\varphi_{kj}(x_j)}{\sum_{k=1}^{K}\prod_{j=1}^{D}\varphi_{kj}(x_j)}.
\end{equation}
where $K$ denotes the number of applied fuzzy rules.

The overall output is obtained by summing the outputs of all fuzzy rules multiplied with a weighted vector:
\begin{equation}\label{defuzzy}
\hat{y} = \sum_{k=1}^{K}\bar{\phi}_k(x)(w_{k0} + \sum_{j=1}^{D}w_{kj}x_j),
\end{equation}

The structure learning process therefore aims to optimize the parameters of the Gaussian membership functions in (\ref{Gaussian_A}) for each fuzzy rule, i.e., $m_{kj}$ and $\sigma_{kj}$, $j\in \{1,\cdots,D\}$, which is done through the FCM clustering method\cite{chiu1994fuzzy}. In turn, the goal of the parameter learning process is to identify the output weights $w_{k0},\cdots,w_{kD}$ in (\ref{defuzzy}), done with a least-squares algorithm\cite{de2009sofmls}. A more detailed description of the structure learning procedure follows next.

\subsection{Fuzzy C-Means for the structure learning}
As mentioned, the structure learning process is governed by the FCM algorithm. The fuzzy rules are generated by clustering training data into several groups where each group corresponds to a fuzzy rule.
This procedure follows.

Let $\clD:=\{(X_i,Y_i)|X_i\in \mathbb{R}^D, Y_i\in \mathbb{R},\ i\in\{1,\cdots,N\} \}$ denote the training data set and $x_{ij}$ denote the $j$-th feature of the $i$-th sample $X_i$. The aim of FCM algorithm is to partition $N$ samples into $K$ groups $\mathcal{C} :=\{\mathcal{C}_1,\cdots,\mathcal{C}_K\}$, where $K$ denotes the total number of fuzzy rules and is often assumed to be known as a priori. The most important task for the structure learning is to identifying the total number $K$ of clusters, where the $k-$th center can be identified follows:
\begin{equation}\label{cluster1}
 \bm_k = \mbox{arg}\ds\min_{\bm_k} \frac{1}{2} \sum_{k=1}^{K}\sum_{i=1}^{|\clC_k|}u_{ik}^{\alpha}||X_i-\bm_k||^2
\end{equation}
where $|\clC_k|$ denotes the $k-$th cluster, and $\alpha \geq 1$ determines the level of cluster fuzziness. ($\alpha$ is commonly set to 2.) With an iterative technique, the FCM algorithm then assembles and refines the clusters. All the $K$ centers at iteration $t=0$ are initialized randomly,
i.e., $\{\bm_1(0),\cdots,\bm_K(0)\}$, the procedures from $t+1$ iterates as follows:
\begin{itemize}
  \item Update the membership value:
        \begin{equation}\label{center_c}
          u_{ik}^{\alpha}(t+1) = \frac{1}{\sum_{c=1}^{K}(\frac{||X_i-\bm_k(t)||}{||X_i-\bm_c(t)||})^{\frac{2}{\alpha-1}}}
        \end{equation}
  \item Update the cluster center:
        \begin{equation}\label{member_c}
          \bm_k(t+1) = \frac{\sum_{i=1}^{N}u_{ik}^{\alpha}(t+1)X_i}{\sum_{i=1}^{N}u_{ij}^{\alpha}(t+1)}
        \end{equation}
\end{itemize}
The algorithm stops to optimize when the values of centers stay unchanged during the iteration, and the obtained centers become the centers of each fuzzy set. Meanwhile, the standard variance $\sigma_{kj}$ of $k-$th fuzzy set can be written as
\begin{equation}\label{standard_v}
\sigma_{kj} = \sqrt{ \ds\sum_{i=1}^{N}u_{ik}^{\alpha}(X_{ij}-m_{kj})^2/\ds\sum_{i=1}^{N}u_{ik}^{\alpha}}
\end{equation}
And voilà, we have all the parameters in the antecedent layer.

To calculate the output weights in the consequent layer, parameter learning is described next with a closed-form solution.

\subsection{Closed-form solution for the parameter learning}
Parameter learning begins with a hidden matrix defined as $H(X) := [H_1,\cdots, H_K]\in\mathbb{R}^{N\times K(D+1)}$, where
\begin{equation}\label{hiddenH}
H_k(X)=\begin{bmatrix}
\bar{\phi}_k(X_1) & \bar{\phi}_k(X_1)x_{11} & \cdots & \bar{\phi}_k(X_1)x_{1D} \\
\bar{\phi}_k(X_2) & \bar{\phi}_k(X_2)x_{21} & \cdots & \bar{\phi}_k(X_2)x_{2D} \\
\vdots & \vdots & \ddots & \vdots \\
\bar{\phi}_k(X_N) & \bar{\phi}_k(X_N)x_{N1} & \cdots & \bar{\phi}_k(X_N)x_{ND},
\end{bmatrix}
\end{equation}
and the output vector $Y:=[Y_1,\cdots,Y_N]$.

The output weight matrix $\bw\in\mathbb{R}^{K(D+1)}$ can be written as:
\begin{equation}\label{weight2}
  \bw = [w_{10},\cdots,w_{1d},\cdots,w_{K0},\cdots,w_{KD}]^T,
\end{equation}
it can be identified by calculating the optimization problem follows,
\begin{equation}\label{weight1}
  \ds\min \mathcal{F}_s(\bw;X) = \ds\min_{\bw} \frac{1}{2}||Y-f(\bw;X)||^2 + \frac{\mu}{2}||\bw||^2,
\end{equation}
where $f(\bw;X) = H(X)\bw$, $\mu>0$ trades off the model performance between training error and model generalization.
(\ref{weight1}) is a standard least-squares optimization problem, which can be obtained with a closed form solution follows:
\begin{equation}\label{weight3}
  \bw = (H^T H + \mu I)^{-1}H^T Y,
\end{equation}
where $I$ is the identity matrix with a dimension of $K(D+1)$.

\subsection{Semi-supervised fuzzy regression with ICR}
As mentioned before, using ICR to solve semi-supervised fuzzy regression problems can push decision boundaries into low-density areas, leading to better generalization performance in semi-supervised scenarios \cite{zhang2017mixup,verma2019interpolation}. We followed the data augmentation techniques in \cite{zhang2017mixup,verma2019interpolation} to generate interpolation consistency loss and involved it into the objective  function (\ref{weight1}).

The training set consists of a labeled data set $X\in \mathcal C_s$ and an unlabeled data set $U\in \mathcal C_u$, denoted as $\mathcal C = \mathcal C_s \cup \mathcal C_u$. Accordingly, the number of the training samples can be expressed as $N = N_s + N_u$. ICR augments this training set with virtual samples constructed from the unlabeled data set $\mathcal C_u$ as follows:
\begin{eqnarray}
  \tilde{U} &=& \lambda U_1 + (1-\lambda) U_2, \ U_1, U_2\in \mathcal C_u, \label{interpolation}  \\
  f(\bw; \tilde{U}) &=& \lambda f(\bw; U_1) + (1-\lambda) f(\bw; U_2)
\end{eqnarray}
where $\lambda$ is randomly sampled from a beta distribution.

The objective function of semi-supervised fuzzy regression (SFR) with ICR is
\begin{equation}\label{obj_all}
 \ds\min \mathcal F(\bw;X,U) = \ds\min_{\bw}\mathcal F_s(\bw;X) + \gamma \mathcal F_u(\bw;U),
\end{equation}
where
\begin{eqnarray}\label{obj_semi}
 \mathcal F_u(\bw;U) &=& ||f(\bw;\tilde{U}) - (\lambda f(\bw;U_1)+(1-\lambda)f(\bw;U_2))||^2\nonumber\\
 &=& ||H(\tilde{U})\bw - (\lambda H(U_1)\bw + (1-\lambda) H(U_2)\bw)||^2, \nonumber\\
 &\triangleq& ||B(U)\bw||^2,
\end{eqnarray}
where $B(U) = H(\tilde{U})- \lambda H(U_1) - (1-\lambda)H(U_2)$.
The closed-form solution of (\ref{obj_all}) is:
\begin{eqnarray}
    \mathcal \bw = (\gamma B^T(U)B(U) + \mu I + H^T(X)H(X))^{-1}H^T(X)Y.
\end{eqnarray}

\section{Distributed semi-supervised fuzzy regression with ICR}

In this section, we extend the CSFR model to its distributed version and design distributed training algorithms for it. Since the CSFR model is sequentially trained by structure learning and parameter learning, we will develop a distributed structure learning and a distributed parameter learning sequentially.

Our distributed computing scenario is conceived as an undirected graph $\mathcal{G}=\{\clL,\xi\}$, of $L$ agents (nodes) connected by $E$ edges, where $\clL$ and $\xi$ denotes the nodes set and the edges set, respectively. Agent $l$ is the target agent, and $\clN_l$ is the set of agents neighboring agent $l$. Given a dataset $\clD:=\{(X_i,Y_i)| i\in \clN \}$, let $\{\clD^1,\cdots,\clD^L\}$ be its decomposition in entirety. The subset $\clD^l$ is the data housed on the $l-$th node, where $l\in\clL$.  The subset of samples located on the $l-$th node can be then denoted as $\clC^l$. Within each subset $\clC^l$, $\clC^l_k$ denotes the subset of smples in the $k-th$ cluster of agent $l$ such that $\bigcup_{k=1}^K \clC^l_k = \clC^l$. An illustration of this architecture is shown in Fig.1.
\begin{figure*}[!htbp]
  \centering
  \includegraphics[width=1.4\columnwidth]{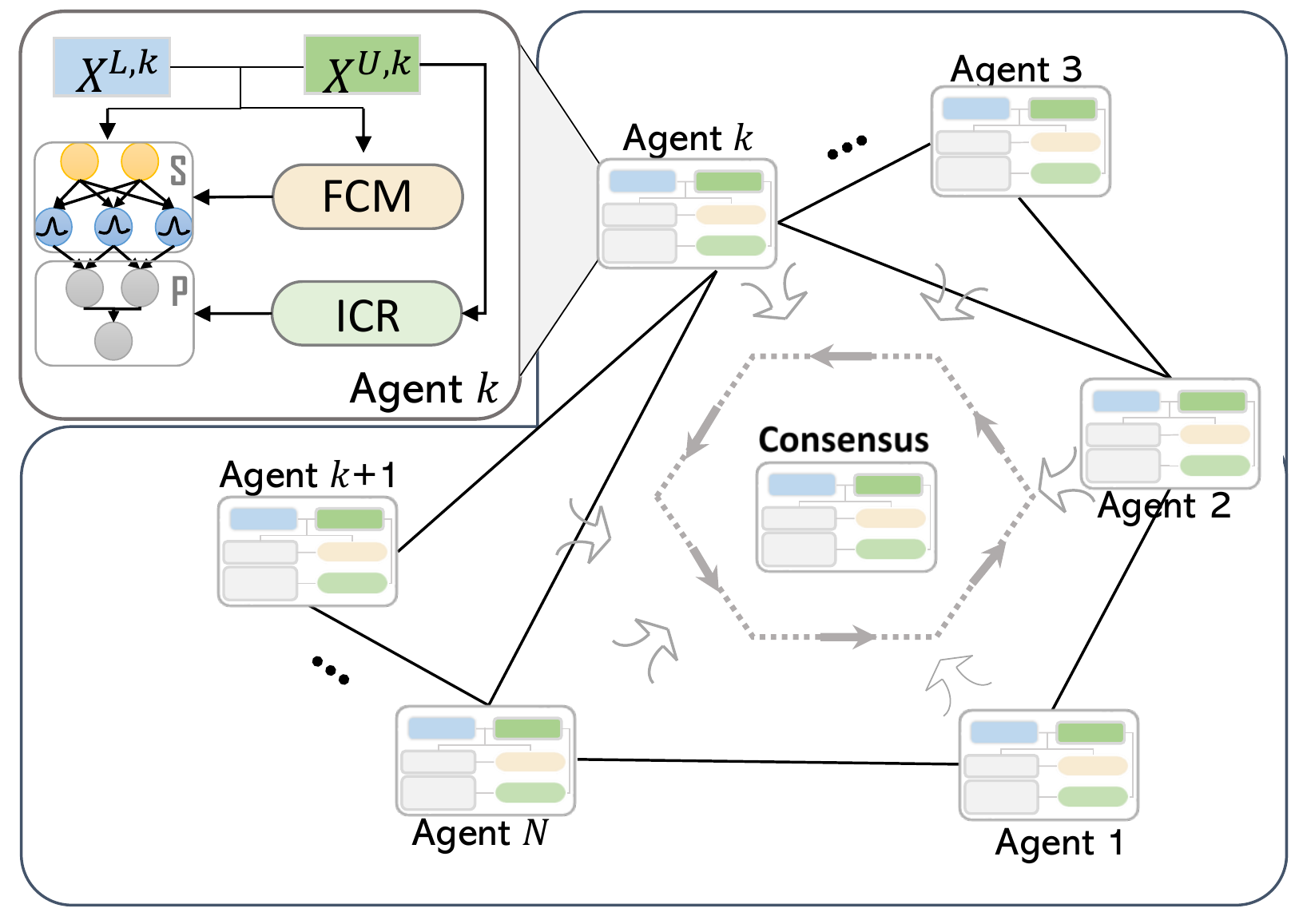}
  \vspace{-10pt}
  \caption{\normalsize Architecture of the DSFR model. The upper-left part depicts detailed structures of each local model, which is presented in the bottom-right part. Different colors are used to distinguish different types of data and methods.}
  \label{fig:architecture}
\end{figure*}

The corresponding structure learning problem is solved through a consensus strategy, formulated as
\begin{subequations}\label{str}
\begin{eqnarray}
   \ds\min_{\bm^l_k} \frac{1}{2} \sum_{l=1}^{L}\sum_{k=1}^{K}\sum_{X^l_i\in \clC^l_k}(u_{ik}^q)^{\alpha}||X^l_i-\bm^l_k||^2\label{str_a}\\
   \mbox{s.t.}\quad  \bm^l_k = \br_k, \ l\in \clL,\ k\in\clK, \label{str_b}
\end{eqnarray}
\end{subequations}
where $\bm^l_k$ represents the local center of the $k-h$ fuzzy set on agent $l$, $\br_k$ denotes the global center, which integrates all the local centers, and $|\clC^l_k|$ is the cardinality operation for a local subset. Moreover, (\ref{str_b}) is the constraint that assures all local centers coincide at one  global center. Notably, all the local variables among different agents can be parallelly computed, thus improving the computing speed.

After Identifying all the global centers, the global standard variance can be calculated via
\begin{equation}\label{glo_sta}
  \bar{\sigma}_{kj} = \sqrt{\frac{1}{N}\ds\sum_{l=1}^{L}|\clC^l|(\sigma^l_{kj})^2}
\end{equation}
where $|\clC^l|$ is the cardinality of local subset $\clC^l$ on agent $l$,  $\sigma^l_{kj}$ denotes the $j$-th element of fuzzy rule $k$'s standard variance on subset $\clC^l$, and $\bar{\sigma}_{kj}$ is the corresponding to global standard variance for all agents. {\color{black} Note that the standard variance $\sigma^l_{kj}$ corresponding to the $k$-th rule and $j$-th dimension is calculated in the following element-wise method:
\begin{equation}\label{compo_sta}
  \sigma^l_{kj} = \sqrt{\ds\sum_{X_i^l\in\clC^l_k}(u_{ik}^l)^{\alpha}(X_{ij}^l-\bm_{kj}^l)^2/ \ds\sum_{X_i^l\in\clC^l_k}(u_{ik}^l)^{\alpha}}
\end{equation}
where $X_{ij}^l$ and $\bm_{kj}^l$ are the $j$-th components of $X_{i}^l$ and $\bm_{k}^l$, respectively.
}

The parameter learning process can be similarly modeled as follows:
\begin{subequations}\label{par}
\begin{eqnarray}
   \ds\min_{\bw^l} &\frac{1}{2}\sum_{l=1}^{L}(||Y^l-H(X^l)\bw^l||^2 \nonumber\\& + \gamma ||B(U^l)\bw^l||^2) + \frac{\mu}{2} ||\bz||^2, \label{par_a}\\&
   \mbox{s.t.}\quad  \bw^l = \bz, \ l\in\clL, \label{par_b}
\end{eqnarray}
\end{subequations}
where $\bw^l$ represents $l-$th agent's the local output weight, and $\bz$ is a the global weight that integrates local weights for all agents.

\subsection{Distributed FCM}
The optimization problem (\ref{str}) is nonconvex, so using an exhaustive search method to solve the problem would not be efficient. Further, centralized clustering methods, such as the FCM algorithm used for structural learning, have several shortcomings including high communications overheads, no  attention to privacy, and poor scalability when coping with large-scale data. Hence, a distributed variant of the centralized FCM algorithm is needed to address these issues. Accordingly, our formulation of a DFCM algorithm follows.

The first step is to construct the following augmented Lagrangian for (\ref{str}):
\begin{eqnarray}\label{aug_s}
 \clL_s(\bm,\br,\boldsymbol{\lambda}_s) =  && \frac{1}{2} \sum_{l=1}^{L}\sum_{k=1}^{K}\sum_{X^l_i\in \mathcal{C}_k^l}(u_{ik}^q)^{\alpha}||X^l_i-\bm^l_k||^2 \nonumber\\ && +  \ds\sum_{l=1}^{L}\sum_{k=1}^{K}\boldsymbol{\lambda}_{s,kl}^T(\bm^l_k - \br_k)  \nonumber\\ && +
  \frac{1}{2}\rho_{s} \ds\sum_{l=1}^{L}\sum_{k=1}^{K}||\bm^l_k - \br_k||^2
\end{eqnarray}
where $\boldsymbol{\lambda}_{s,kl}$ denotes the Lagrange multiplier and $\rho_{s}$ is a positive penalty factor.
Denoting these parameters at iteration $t$ as $\br(t)$ and $\boldsymbol{\lambda}_s(t)$,
the variables are obtained iteratively using the following ADMM-based updating steps:
\begin{eqnarray}
(u_{ik}^l)^{\alpha}(t+1) = \frac{1}{\sum_{c=1}^{K}(\frac{||X_i^l-\bm_k^l(t)||}{||X_i^l-\bm_c^l(t)||})^{\frac{2}{\alpha-1}}},\label{ADMM_u}\\
\bm^l(t+1) = \mbox{arg}\ds\min_{\bm}\clL(\bm^l,\br(t),\boldsymbol{\lambda}_s(t)),\label{ADMM_s1}\\
\br(t+1) = \mbox{arg}\ds\min_{\br}\clL(\bm^l(t+1),\br,\boldsymbol{\lambda}_s(t)),\label{ADMM_s2}\\
\boldsymbol{\lambda}_{s,kl}(t+1) = \boldsymbol{\lambda}_{s,kl}(t) + \rho_{s}(\bm^l_k(t+1) - \br_k(t+1)).\label{ADMM_s3}
\end{eqnarray}
It is worth noting that (\ref{ADMM_s1}) can be solved in parallel among different agents.

The cluster centers $\bm^l(t+1)$ for each agent $l$ are updated through the same assignment and update steps from the centralized FCM algorithm. But, it's easy to know that the solutions (\ref{ADMM_s1}) and (\ref{ADMM_s2}) can be calculated by setting the partial derivative of the corresponding variable to zero. Thus, the closed-form solution of (\ref{ADMM_s1}) and (\ref{ADMM_s2}) are as follows:

\begin{equation}\label{closed-form-m}
  \bm_k^l(t+1) = \frac{\sum_{X^l_i\in \mathcal{C}_k^l}(u_{ik}^l(t))^{\alpha}X_i-\boldsymbol{\lambda}_{s,kl}(t)+\rho_{s}\br_k(t)}{\sum_{X^l_i\in \mathcal{C}_k^l}(u_{ik}^l(t))^{\alpha}+\rho_{s}}
\end{equation}

\begin{eqnarray}\label{close_r}
\br_k(t+1) = \frac{1}{\rho_{s}}\bar{\boldsymbol{\lambda}}_{s,kl}(t) + \bar{\bm}^l_k(t+1),
\end{eqnarray}
where
\begin{eqnarray}\bar{\bm}^l_k(t+1) =\frac{1}{L} \sum_{l=1}^{L}\bm^l_k(t+1),\label{bar_m}\\  \bar{\boldsymbol{\lambda}}_{s,kl}(t) = \frac{1}{L} \sum_{l=1}^{L}\boldsymbol{\lambda}_{s,kl}(t). \label{bar_lambda}
\end{eqnarray}

Algorithm \ref{alg1} summarizes the above approach. The convergence behavior is examined by checking the norms of the following two residuals:
\begin{eqnarray}\label{converg_alg1}
||\bm^l_k(t) - \bm^q_k(t)||^2 \leq \epsilon_1,\\
||\boldsymbol{\lambda}_{s,k}^{l}(t) - \boldsymbol{\lambda}_{s,k}^{l}(t-1)||^2 \leq \epsilon_2.
\end{eqnarray}
\begin{algorithm}
\caption{Distributed FCM (\ref{str})} \label{alg1}
\begin{algorithmic}
\STATE{\textbf{Initialization:} Set $t=0$ and the Lagrange multipliers $\boldsymbol{\lambda}_{s,kl}(t)=\mathbf{0}$. Initialize cluster centers and assign them to each agent $l$.}
\FOR{$t=0,1,2,\cdots,$}
\STATE {\textbf{Update the membership value $(u_{ik}^l)^{\alpha}$} by (\ref{ADMM_u})} for each agent $l$.
\STATE {\textbf{Update the local variables $\bm^l(t+1)$} by (\ref{closed-form-m})} and broadcast it to each agent $l$.
\STATE {\textbf{Update the global variables $\br(t+1)$} by  (\ref{close_r}) and broadcast them to each agent $l$. }
\STATE {\textbf{Update the dual variables $\boldsymbol{\lambda}_s(t+1)$} by (\ref{ADMM_s3}) and broadcast them to each agent $l$}
\ENDFOR
\end{algorithmic}
\end{algorithm}

The computation complexity of DFCM relies on the local variables update and global variables update. The computation complexity for DFCM is $\mathcal{O}(NDK^2T_1)$, where $T_1$ is the required iteration of DFCM. Note that the DFCM is built on the well-known ADMM algorithm, which generally converges to modest accuracy (such as $\epsilon<10^{-3}$) within a few tens of iterations. Therefore, the total computational complexity of DFCM is quite limited.

\subsection{Distributed ICR}
Likewise, a distributed variant of ICR  is needed to tackle the regression problem.  In  contrast  to  existing  DDSL algorithms \cite{xie2019distributed,xie2019distributedelm}, which are typically based on graph operations, DICR  is more faster and more accurate with dramatically reduced computing times, especially with large-scale datasets. The formulation follows.

The augmented Lagrangian for (\ref{par}) is
\begin{eqnarray}\label{aug_p}
   \mathcal L(\bw^l, \bz, \boldsymbol{\beta_q})= &&\frac{1}{2}\sum_{l=1}^{L}(||Y^l-H(X^l)\bw^l||^2 + \gamma ||B(U^l)\bw^l||^2) \nonumber\\
   &&+ \frac{\mu}{2} ||\bz||^2
  + \sum_{l=1}^{L}\boldsymbol{\beta}_{l}^T(\bw^l - z)
    \nonumber\\&&+ \frac{1}{2}\rho_p\sum_{l=1}^{L}||\bw^l - z||^2
\end{eqnarray}
where $\boldsymbol{\beta}_{l}$ is the Lagrange multiplier, and $\rho_p$ is a small positive penalty parameter. This augment Lagrangian is then solved using ADMM method by

\begin{eqnarray}
\bw^l(t+1) = \mbox{arg}\min_{\bw}\mathcal L(\bw,\bz(t),\boldsymbol{\beta}(t)),\label{ADMM_p1}\\
\bz^{l}(t+1) = \mbox{arg}\min_{\bz}\mathcal L(\bw(t+1),\bz,\boldsymbol{\beta}(t)),\label{ADMM_p2}\\
\boldsymbol{\beta}_{l}(t+1) = \boldsymbol{\beta}_{l}(t) + \rho_p(\bw^l(t+1) - \bz(t+1)).\label{ADMM_p3}
\end{eqnarray}
The closed-form solution of (\ref{ADMM_p1}) is clear;  $w$ on the $l$-th agent is updated by
\begin{equation}\label{update_mix_up_semi_w}
 \bw^l(t+1) = Q((H^l(X))^T Y^l + \rho_p\bz(t)-\boldsymbol{\beta_l}(t)),
\end{equation}
where $I$ is the identity matrix with dimension $K(D + 1)$ and $Q$ can be obtained by
\begin{equation}
 Q = ((H^l(X))^T H^l(X) + \gamma B^T(U)B(U) + \mu I)^{-1}.
\end{equation}

The solution for (\ref{ADMM_p2}) is similarly found by
\begin{equation}\label{update_mix_up_semi_z}
 \bz(t+1) = \frac{\sum_{l=1}^{L}(\beta_{l}+\rho_p \bw^l(t+1))}{\mu+\rho_p L},
\end{equation}

\begin{algorithm}
\caption{Distributed ICR \label{alg_mix_up_semi}}
\begin{algorithmic}
\STATE{\textbf{Initialization:} set $t=0$ and initialize the global weight $\bz(t)$ and Lagrange multipliers $\beta(t)$ for each agent $l$.}
\FOR{$t=0,1,2,\cdots,$}
\STATE {\textbf{Augmented data generation}}:\\
Randomly select $M$ samples from the unlabeled data set $\mathcal{C}_u$ as $U_1$ and another $M$ samples as $U_2$. Then generate $M$ number of $\lambda$ using a beta distribution, and generate $M$ augmented samples using the interpolation technique in (\ref{interpolation}).\\
\STATE {\textbf{Update the local variables $w^l(t+1)$} via (\ref{update_mix_up_semi_w}) for each agent $l$.}
\STATE {\textbf{Update the global variable $z(t+1)$} via  (\ref{update_mix_up_semi_z}) and broadcast it to the other agents. }
\STATE {\textbf{Update the dual variables $\boldsymbol{\beta}(t+1)$} by (\ref{ADMM_p3}) and broadcast them to the other agents.}
\ENDFOR
\end{algorithmic}
\end{algorithm}

The computation complexity of DICR is $\mathcal{O}(KDLT_2)$ if we ignore the cost of data augmentation, where $T_2$ is the iteration number required by ADMM procedure. As we analyzed before, $T_2$ is generally about a few tens. Therefore, the total computation complexity of DICR is small.

\section{Experiments}
To verify the effectiveness of the DSFR framework, we conducted extensive experiments on six different types of datasets. Descriptive statistics are provided in Table \ref{tbl:dataset_detail}; brief descriptions follow:
\begin{itemize}
\item[-] The Airfoil dataset \cite{lopez2008neural}. Assembled by NASA, this dataset contains readings of noise tests with different-sized NACA 0012 airfoils.
\item[-] The WarCraft Master Skill Level (WMSL) dataset \cite{thompson2013video}. A set of playing data drawn from the WarCraft real-time strategy game, this set contains the playing data of 3360 gamers. After removing broken records, the final dataset numbered data on 3338 players.
\item[-] The Combined Cycle Power Plant (CCPP) dataset \cite{tufekci2014prediction}. This set contains 9568 samples of gas and steam turbine loads. We chose the full electrical power load as the target label and used the remaining data to train the model.
\item[-] The California Housing (CH) dataset \cite{pace1997sparse}. A data set of information about real estate in California, we discarded all non-numeric features and treated the median house value of a district as the target.
\item[-] The King County Housing (KCH) dataset \cite{hu2017comparison}. This is a collection of houses sold in King County USA. After removing time-related and restriction-related variables, we used house price as the target and the rest of the features for training.
\item[-] The artificial dataset. We built this dataset to investigate the performance of the DSFR framework. 5,000 samples $X \in \mathbb{R}^8$ were generated from a Gaussian distribution with a mean of $[0, 0.5, \cdots, 3.5]$ and a standard deviation of $[0.2, 0.4, \cdots, 1.6]$. The corresponding label $Y$ was given by: $Y = 0.3\sum_{i=1}^{8}(X_i)^2 + 0.7\sum_{i=1}^{8}(\cos(X_i))$. We denoted the range of labels as $R = max(X) - min(X)$ and generated two types of noise from Gaussian distributions $\mathcal{N}(0.1R,(0.1R)^2)$ and $\mathcal{N}(R,(0.1R)^2)$, respectively. Then, $5\%$ level of type-1 noise and $10\%$ type-2 noise were randomly added to labels.
\end{itemize}

It should be noted that all feature values of these datasets are normalized between -1 and 1. As listed in Table \ref{tbl:dataset_detail}, a 5-fold cross-validation is used in our experiment to randomly generate 5 test sets and 5 training sets, among which we randomly choose 50 samples as labeled data and the rest as unlabeled data. Then we re-conduct the experiment 10 times and get the average results and its standard variance. For fairness, we arrange the training sets on an interconnected networks with five fully connected agents, and we set rule number as 5 in all agents for all these datasets. All the experiments are conducted on computing nodes with 26 cores of 2.11 GHz and 95 GB memory.

\begin{table}[!ht]
\centering
\caption{Dataset Information}
\begin{tabular}{lccccc}
\hline
Dataset     & Features & Samples & Labeled  & Unlabeled        & Test  \\ \hline
Airfoil      & 5        & 1503      & 50       & 1153             & 300     \\
WMSL         & 18       & 3338      & 50       & 2621             & 667     \\
CCPP        & 4        & 9568      & 50       & 7605             & 1913    \\
CH          & 8        & 20640     & 50       & 16462            & 4128    \\
KCH         & 15       & 21613     & 50       & 17191            & 4322    \\ \hline
\end{tabular}
\label{tbl:dataset_detail}
\end{table}

All feature values in all datasets were normalized between -1 and 1. As noted in Table \ref{tbl:dataset_detail}, we randomly generated 5 test sets and 5 training sets using 5-fold cross-validation, from which we randomly chose 50 samples to serve as the labeled data; the remaining samples were unlabeled data. All experiments were conducted on a computing node comprising 26 CPU cores of 2.11 GHz and a memory  of 95 GB. The evaluation metric was normalized root mean square error (NRMSE), calculated by

\begin{equation}\label{NRMSE}
  \mbox{NRMSE} = \sqrt{\frac{1}{N\hat{\sigma}_Y^2}\sum_{i=1}^{N}(\hat{Y}_i-Y_i)^2},
\end{equation}

\begin{table*}[!ht]
\caption{Performance comparison on each dataset}
\resizebox{\textwidth}{!}{
\begin{tabular}{|c|c|c|c|c|c|c|c|}
\hline
\multirow{3}{*}{Datasets} & \multirow{3}{*}{Algorithms} & \multicolumn{6}{c|}{Performance}                                                                      \\ \cline{3-8}
                         &                            & \multicolumn{3}{c|}{Centralized Methods}               & \multicolumn{3}{c|}{Distributed Methods}                   \\ \cline{3-8}
                         &                            & Test NRMSE    & T-test & Time(s) & Test NRMSE    & T-test & Time(s) \\ \hline
\multirow{5}{*}{Airfoil} & FR                         & 0.6657/0.0188 & 0.0011 & 0.0581 & {\color{red}0.6664/0.0334} & {\color{red}0.0052} & {\color{red}0.3283}  \\ \cline{2-8}
                         & LapWNN                     & 1.3967/0.2478 & 0.0002 & 0.2151 & 1.4428/0.3934 & 0.0000 & 7.1624 \\ \cline{2-8}
                         & s-SFR                      & 2.4290/1.5583 & 0.0419 & 0.1183 & {\color{red}1.6619/0.6212} & {\color{red}0.0124} & {\color{red}0.3123} \\ \cline{2-8}
                         & SFR-G                      & 1.3124/0.3607 & 0.0078 & 0.2066 & {\color{red}1.3578/0.1901} & {\color{red}0.0043} & {\color{red}0.4671} \\ \cline{2-8}
                         & SFR-ICR                    & \textbf{0.7423/0.0288} & -      & 0.1363 & \textbf{0.7506/0.0087} & - & 0.8115 \\ \hline
\multirow{5}{*}{WMSL}    & FR                         & 0.6702/0.0176 & 0.0047 & 0.2811 & {\color{red}0.6770/0.0237} & {\color{red}0.0017} & {\color{red}0.6251} \\ \cline{2-8}
                         & LapWNN                     & \textbf{0.7475/0.0865} & 0.0323 & 27.741 & 0.8734/0.0576 & 0.0110 & 10.678 \\ \cline{2-8}
                         & s-SFR                      & 0.9871/0.0788 & 0.0044 & 0.2953 & {\color{red}0.8634/0.0458} & {\color{red}0.0013} & {\color{red}0.6639} \\ \cline{2-8}
                         & SFR-G                      & 0.9199/0.0474 & 0.0468 & 1.0250 & {\color{red}0.8846/0.0412} & {\color{red}0.0015} & {\color{red}1.0134} \\ \cline{2-8}
                         & SFR-ICR                    & 0.7648/0.0311 & - & 0.5166 & \textbf{0.7588/0.0312} & - & 1.4836 \\ \hline
\multirow{5}{*}{CCPP}    & FR                         & 0.2472/0.0075 & 0.0003 & 0.1173 & {\color{red}0.2472/0.0075} & {\color{red}0.0003} & {\color{red}0.3059} \\ \cline{2-8}
                         & LapWNN                     & 0.7475/0.0865 & 0.0001 & 27.741 & 0.8734/0.0576 & 0.0001 & 10.678 \\ \cline{2-8}
                         & s-SFR                      & 0.3342/0.0528 & 0.0571 & 0.0878 & {\color{red}0.3263/0.0227} & {\color{red}0.0022} & {\color{red}0.2690} \\ \cline{2-8}
                         & SFR-G                      & 0.3187/0.0412 & 0.0811 & 7.6291 & {\color{red}0.3557/0.0820} & {\color{red}0.0091} & {\color{red}1.1656} \\ \cline{2-8}
                         & SFR-ICR                    & \textbf{0.2809/0.0100} & - & 0.1227 & \textbf{0.2856/0.0130} & - & 0.5945 \\ \hline
\multirow{5}{*}{KCH}     & FR                         & 0.5631/0.0080 & 0.0019 & 0.2485 & {\color{red}0.5657/0.0038} & {\color{red}0.0001} & {\color{red}0.8836} \\ \cline{2-8}
                         & LapWNN                     & 0.7262/0.0266 & 0.0969 & 93.464 & 0.9348/0.0291 & 0.0256 & 24.184 \\ \cline{2-8}
                         & s-SFR                      & 0.8063/0.0433 & 0.0041 & 0.1947 & {\color{red}0.7704/0.0788} & {\color{red}0.0103} & {\color{red}0.9245} \\ \cline{2-8}
                         & SFR-G                      & 0.7728/0.0432 & 0.0123 & 89.677 & {\color{red}0.7915/0.0922} & {\color{red}0.0011} & {\color{red}9.9487} \\ \cline{2-8}
                         & SFR-ICR                    & \textbf{0.6952/0.0270} & - & 0.5664 & \textbf{0.6673/0.0211} & - & 1.9435 \\ \hline
\multirow{5}{*}{CH}      & FR                         & 0.5426/0.0180 & 0.0000 & 0.8325 & {\color{red}0.5394/0.0219} & {\color{red}0.0000} & {\color{red}3.6808} \\ \cline{2-8}
                         & LapWNN                     & 1.0023/0.0650 & 0.0223 & 94.025 & 1.0382/0.0294 & 0.0000 & 24.801 \\ \cline{2-8}
                         & s-SFR                      & 0.9387/0.1030 & 0.0012 & 0.6695 & {\color{red}0.9498/0.1210} & {\color{red}0.0457} & {\color{red}2.8925} \\ \cline{2-8}
                         & SFR-G                      & 0.9886/0.0113 & 0.0093 & 53.773 & {\color{red}0.9313/0.0584} & {\color{red}0.0324} & {\color{red}9.7301} \\ \cline{2-8}
                         & SFR-ICR                    & \textbf{0.6875/0.0406} & - & 1.1965 & \textbf{0.7703/0.1217} & - & 5.2945 \\ \hline
\multirow{5}{*}{Artificial} & FR                      & 1.0067/0.0022 & 0.0435 & 0.3371 & {\color{red}1.0069/0.0039} & {\color{red}0.0017} & {\color{red}0.7828} \\ \cline{2-8}
                         & LapWNN                     & 1.2384/0.0824 & 0.0052 & 2.5700 & 1.0891/0.0462 & 0.0034 & 8.2684 \\ \cline{2-8}
                         & s-SFR                      & 1.1590/0.1202 & 0.0027 & 0.2177 & {\color{red}1.2021/0.1627} & {\color{red}0.0015} & {\color{red}0.8537} \\ \cline{2-8}
                         & SFR-G                      & 1.0715/0.0478 & 0.0261 & 1.9241 & {\color{red}1.0653/0.0756} & {\color{red}0.0007} & {\color{red}1.1028} \\ \cline{2-8}
                         & SFR-ICR                    & \textbf{1.0409/0.0339} & - & 0.5266 & \textbf{1.0121/0.0094} & - & 1.4357 \\ \hline
\end{tabular}}
\label{tbl:performance}
\end{table*}

To analyze the convergence of our proposed model and the influence of the rule number and the sample number that is chosen for ICR regularization, we fix the Laplacian parameter $\rho_p$ and $\rho_s$ at 0.1 and 0.1, and the parameters $\mu$, $\gamma$, and $\eta$, which is corresponding to the $L_2$norm regularizer, the mix-up regularizer, and the graph-based regularizer, at 0.1, 0.1, and 0.0001, respectively.

To obtain the best results of our model, all parameters are chosen from an interval of $\{10^{-5},10^{-4},10^{-3},10^{-2},10^{-1},1\}$. Besides, we set the FCM parameter $\alpha$ as 1.1 to increase the diversity of fuzzy rules in all experiments. {\color{red} We set the stopping threshold for DFCM and DICR both as $10^{-4}$.}

\begin{figure*}[!ht]
  \centering
  \subfigure[\normalsize DFCM on Airfoil]{
  \includegraphics[width=0.47\columnwidth]{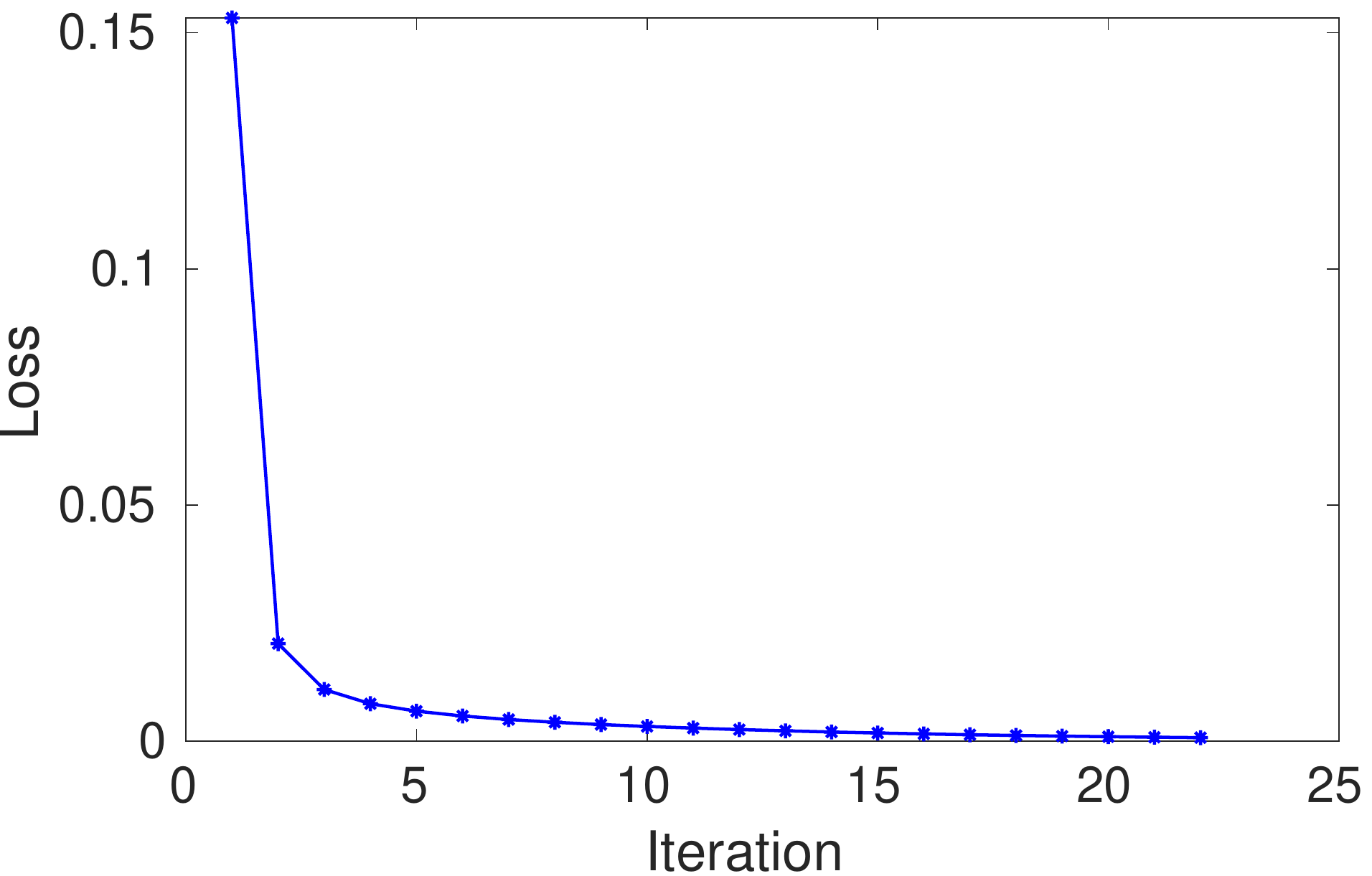}
  }
  \subfigure[\normalsize DFCM on CCPP]{
  \includegraphics[width=0.47\columnwidth]{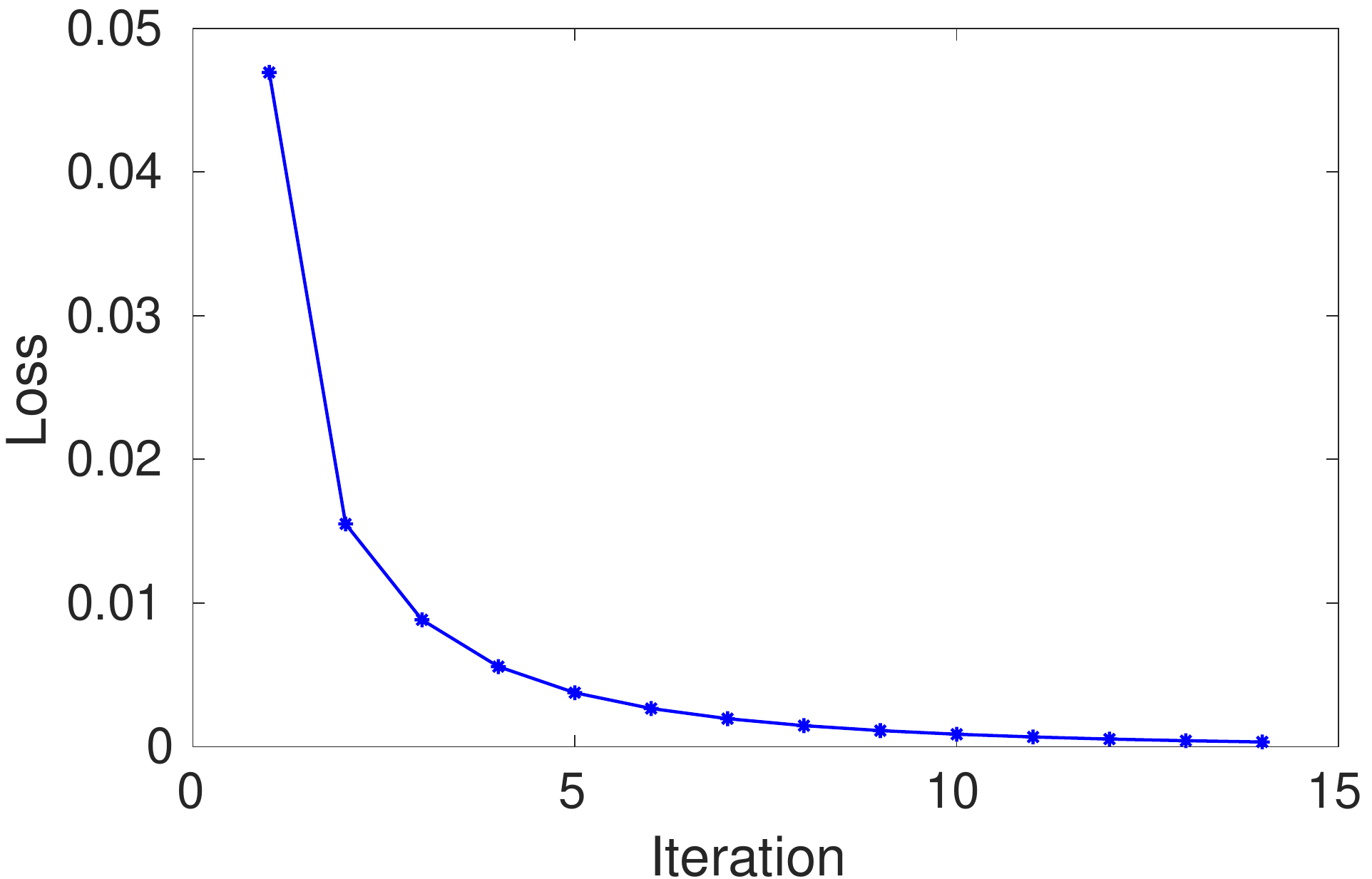}
  }
  \subfigure[\normalsize DFCM on KCH]{
  \includegraphics[width=0.47\columnwidth]{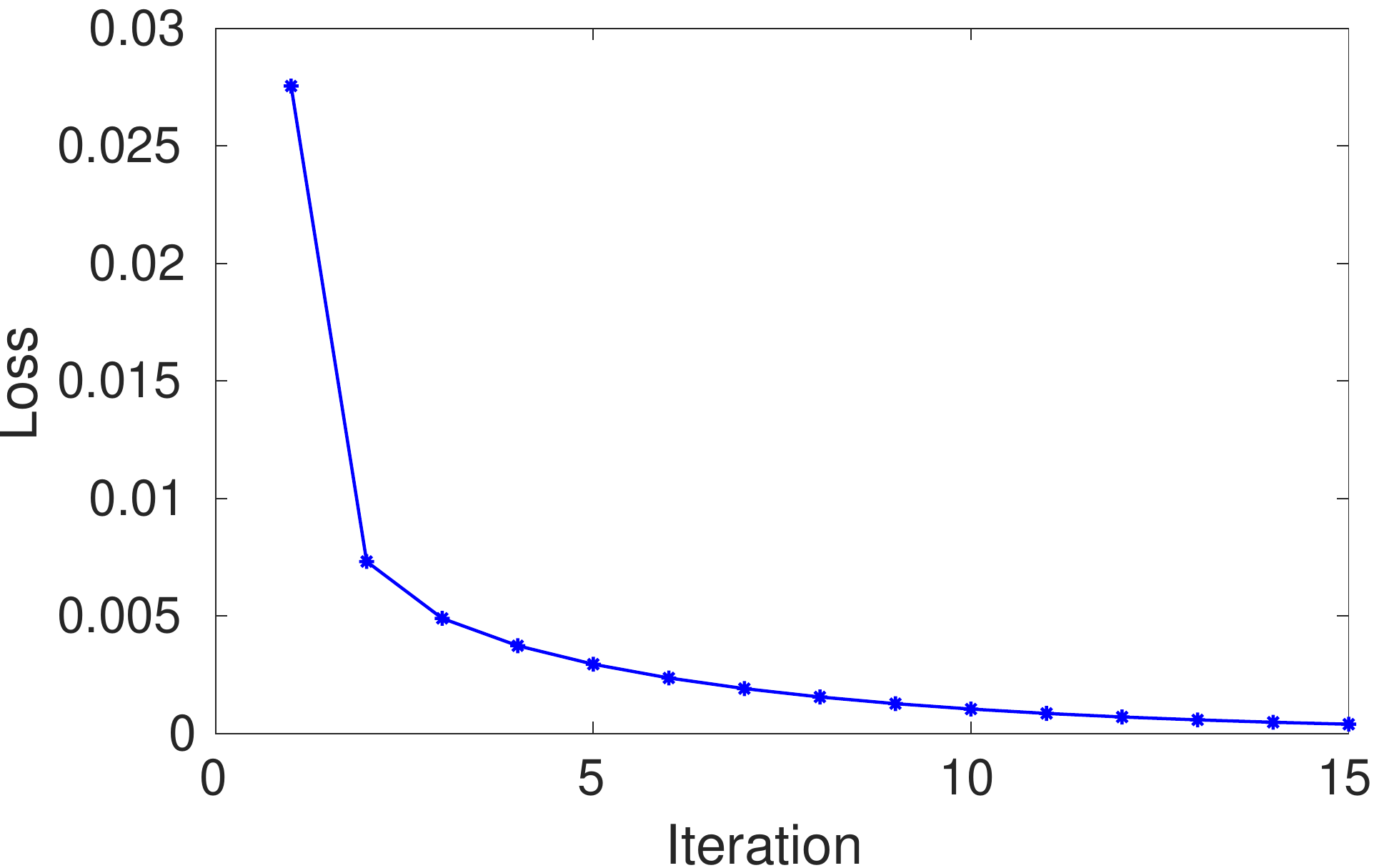}
  }
  \subfigure[\normalsize DFCM on WMSL]{
  \includegraphics[width=0.47\columnwidth]{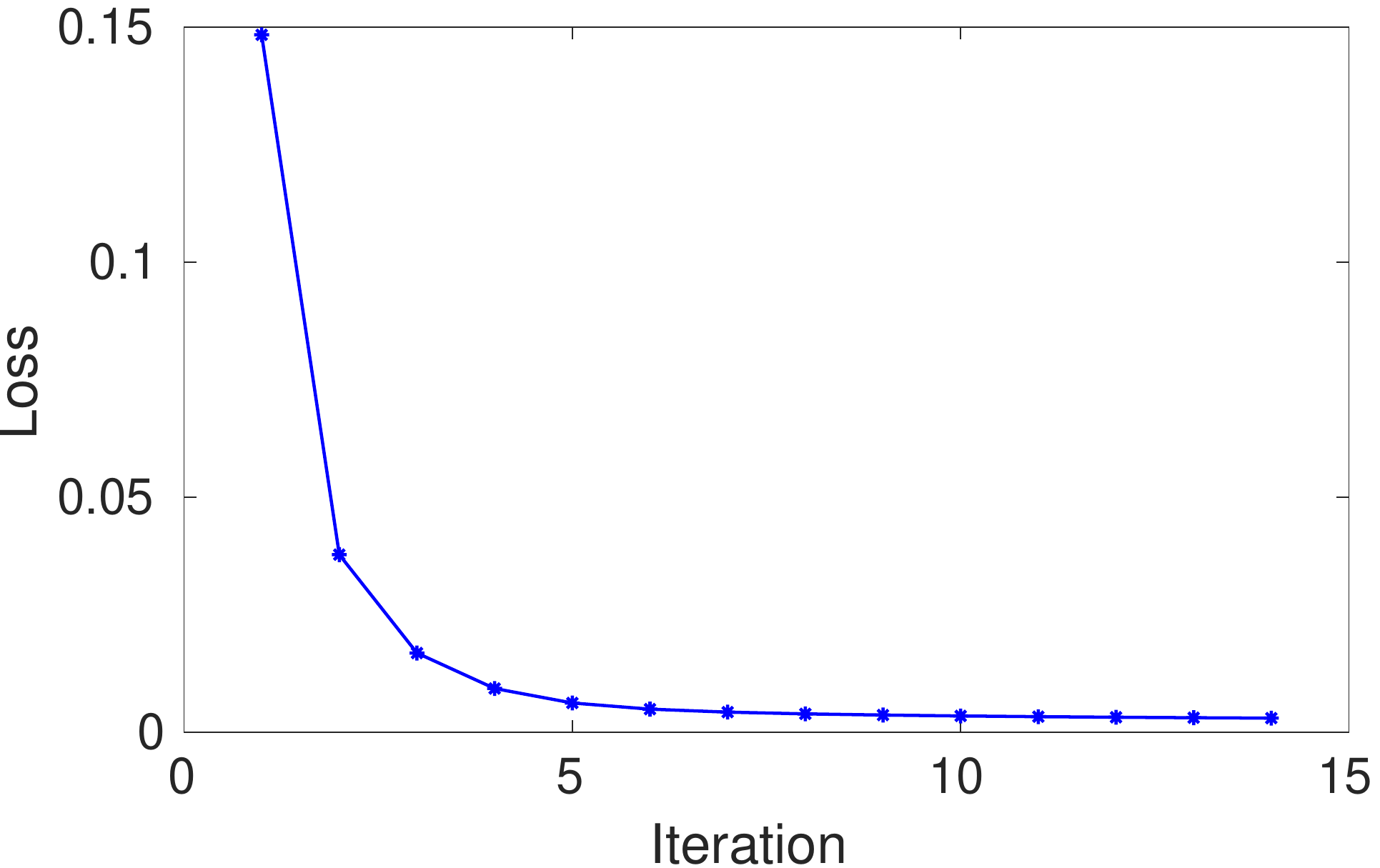}
  }
  \subfigure[\normalsize DICR on Airfoil]{
  \includegraphics[width=0.47\columnwidth]{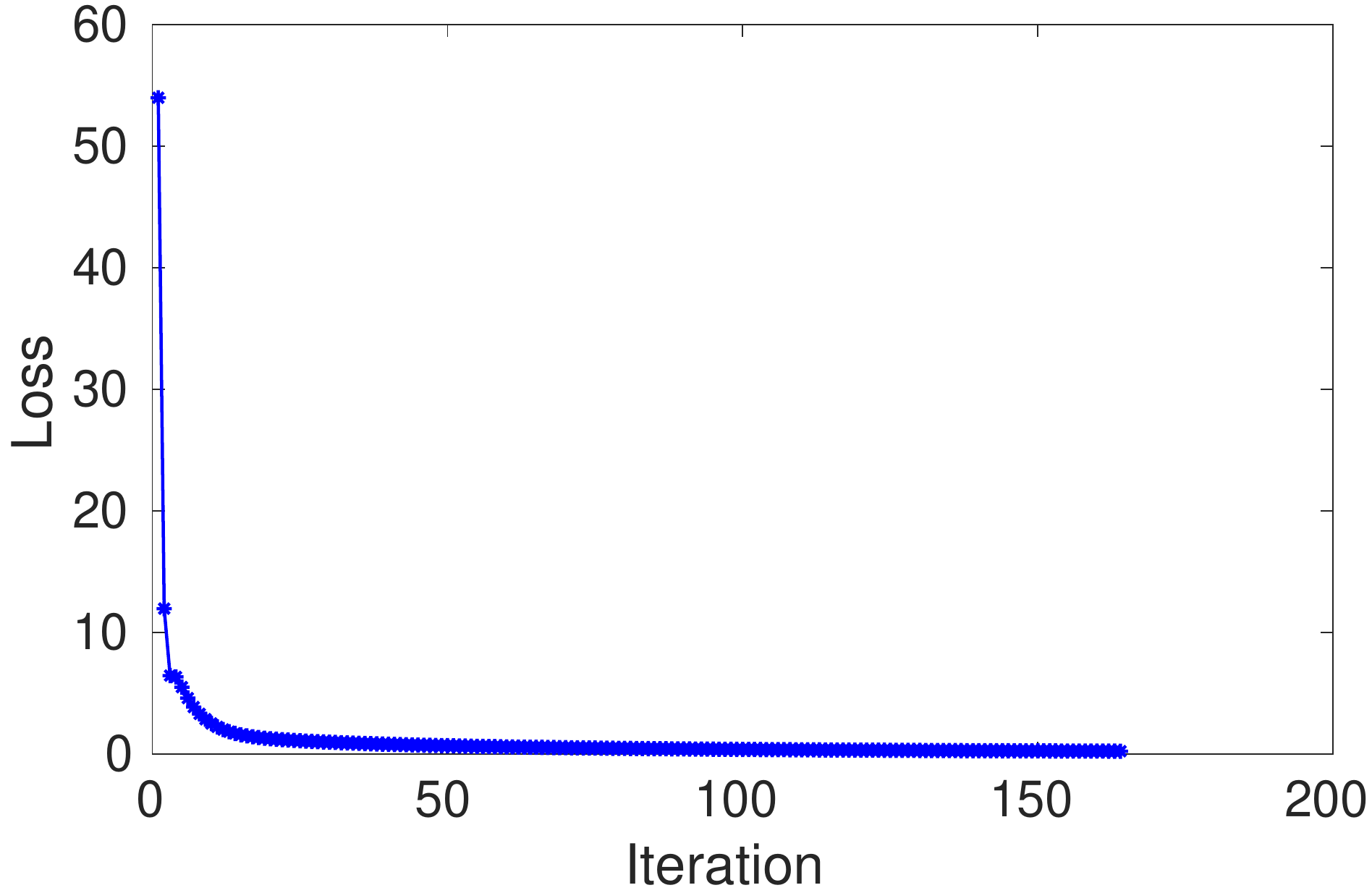}
  }
  \subfigure[\normalsize {\color{red}DICR on CCPP}]{
  \includegraphics[width=0.47\columnwidth]{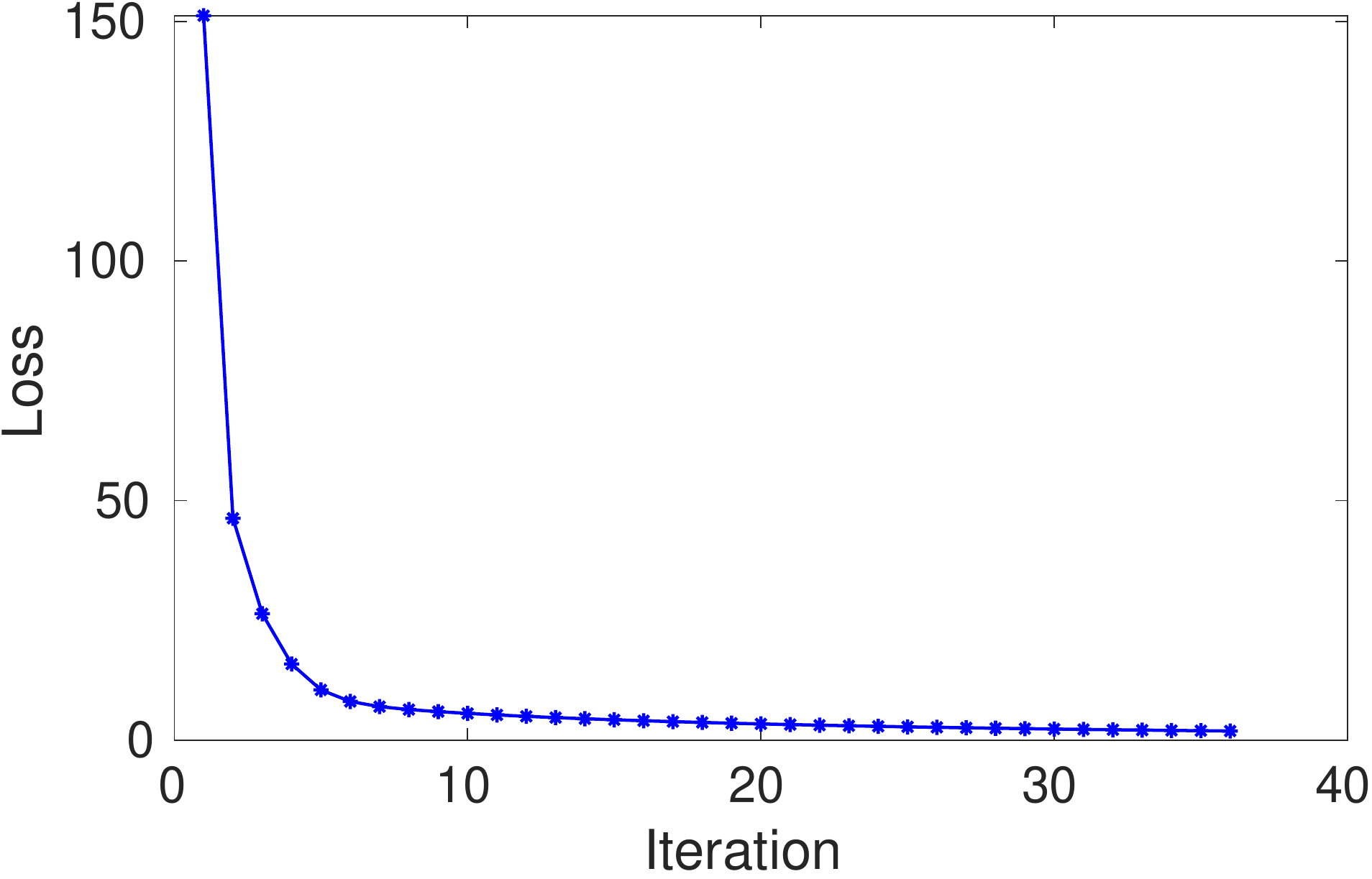}
  }
  \subfigure[\normalsize DICR on KCH]{
  \includegraphics[width=0.47\columnwidth]{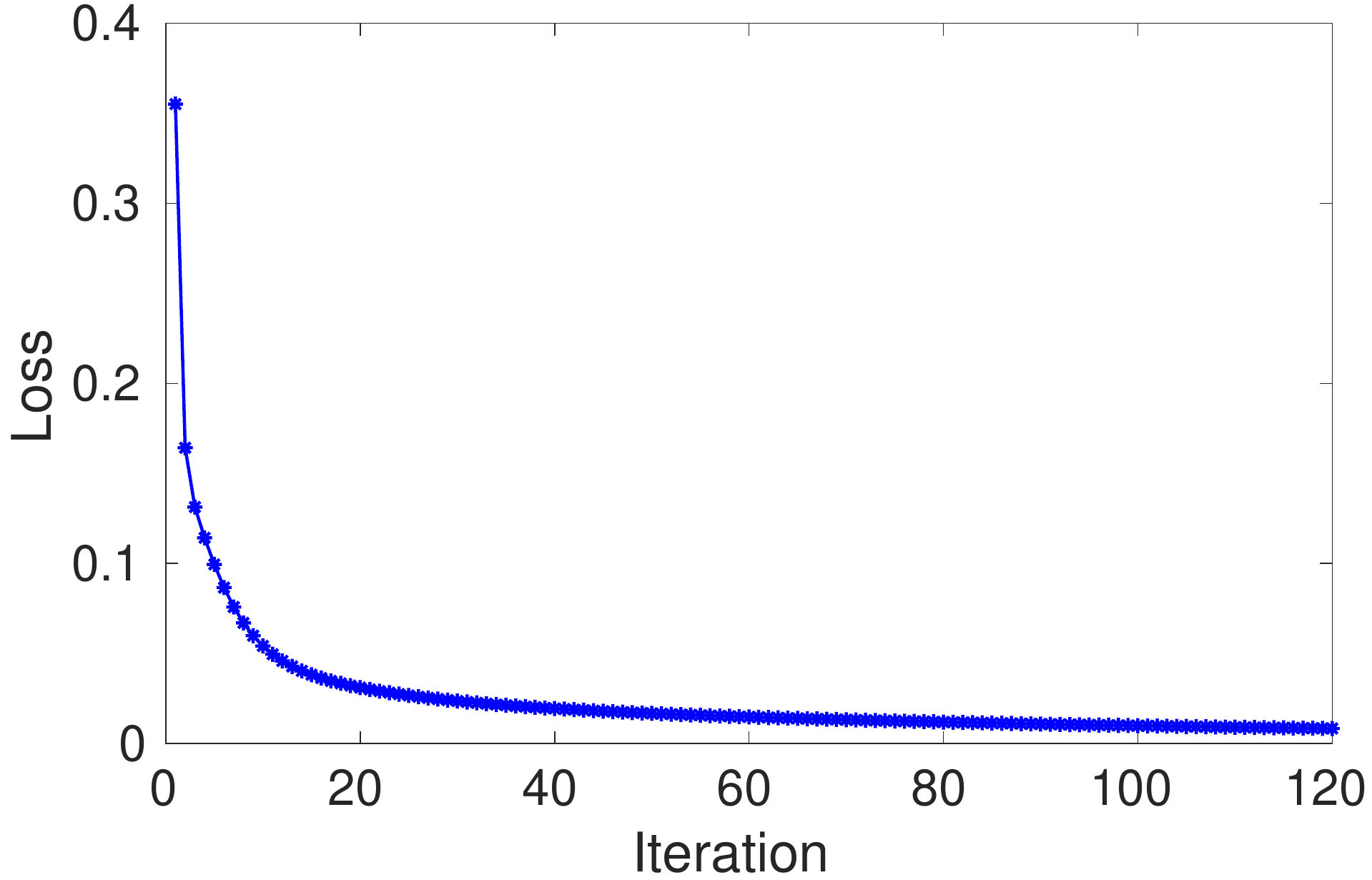}
  }
  \subfigure[\normalsize DICR on WMSL]{
  \includegraphics[width=0.47\columnwidth]{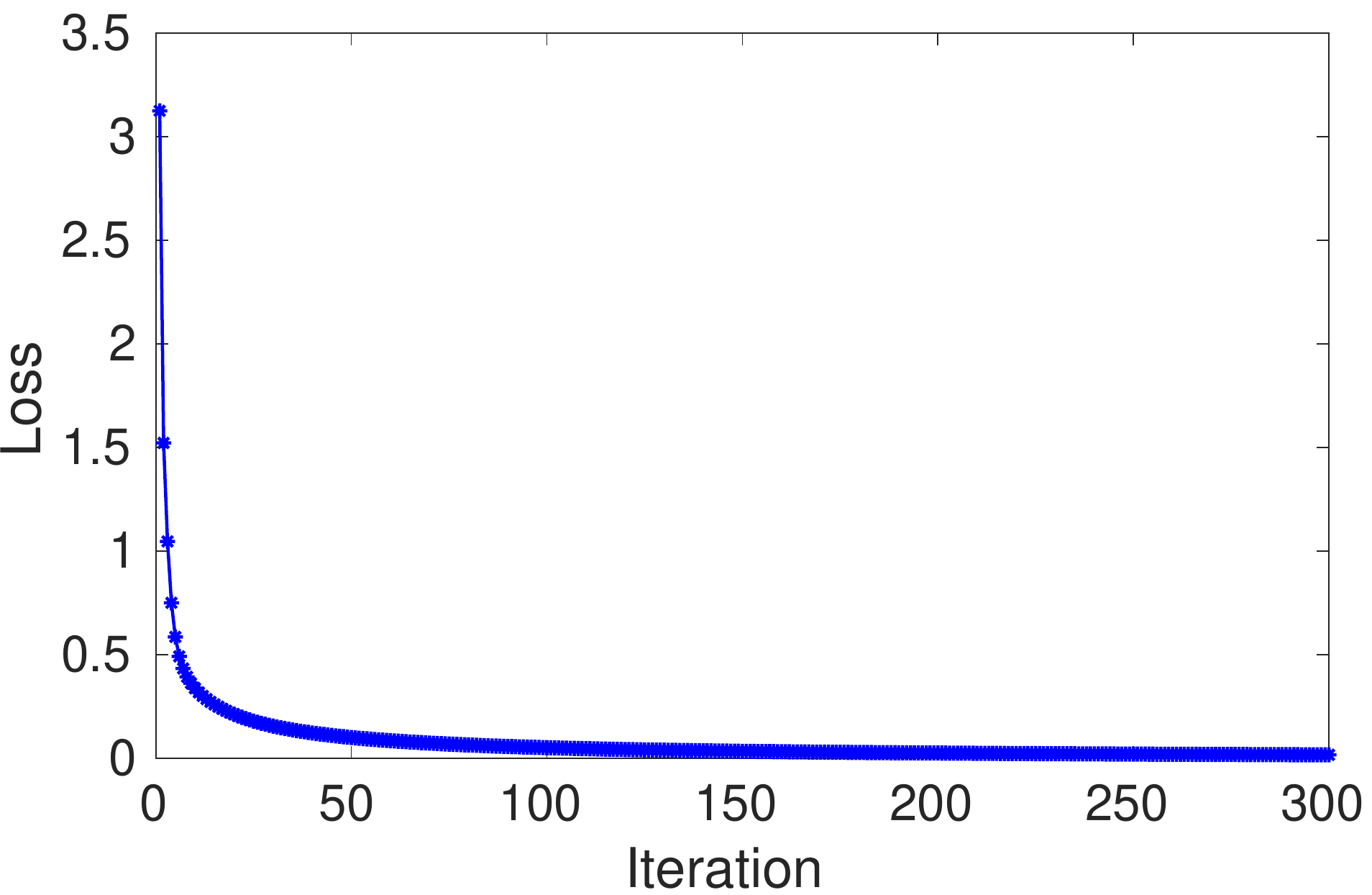}
  }
  \vspace{1pt}
  \centering
  \caption{\normalsize Convergence analysis of the  DFCM method (top) and  DICR method (bottom).}
  \label{fig:conv_ana}
\end{figure*}

\begin{figure*}[!ht]
  \centering
  \subfigure[\normalsize Model parameters on Airfoil]{
  \includegraphics[width=0.47\columnwidth]{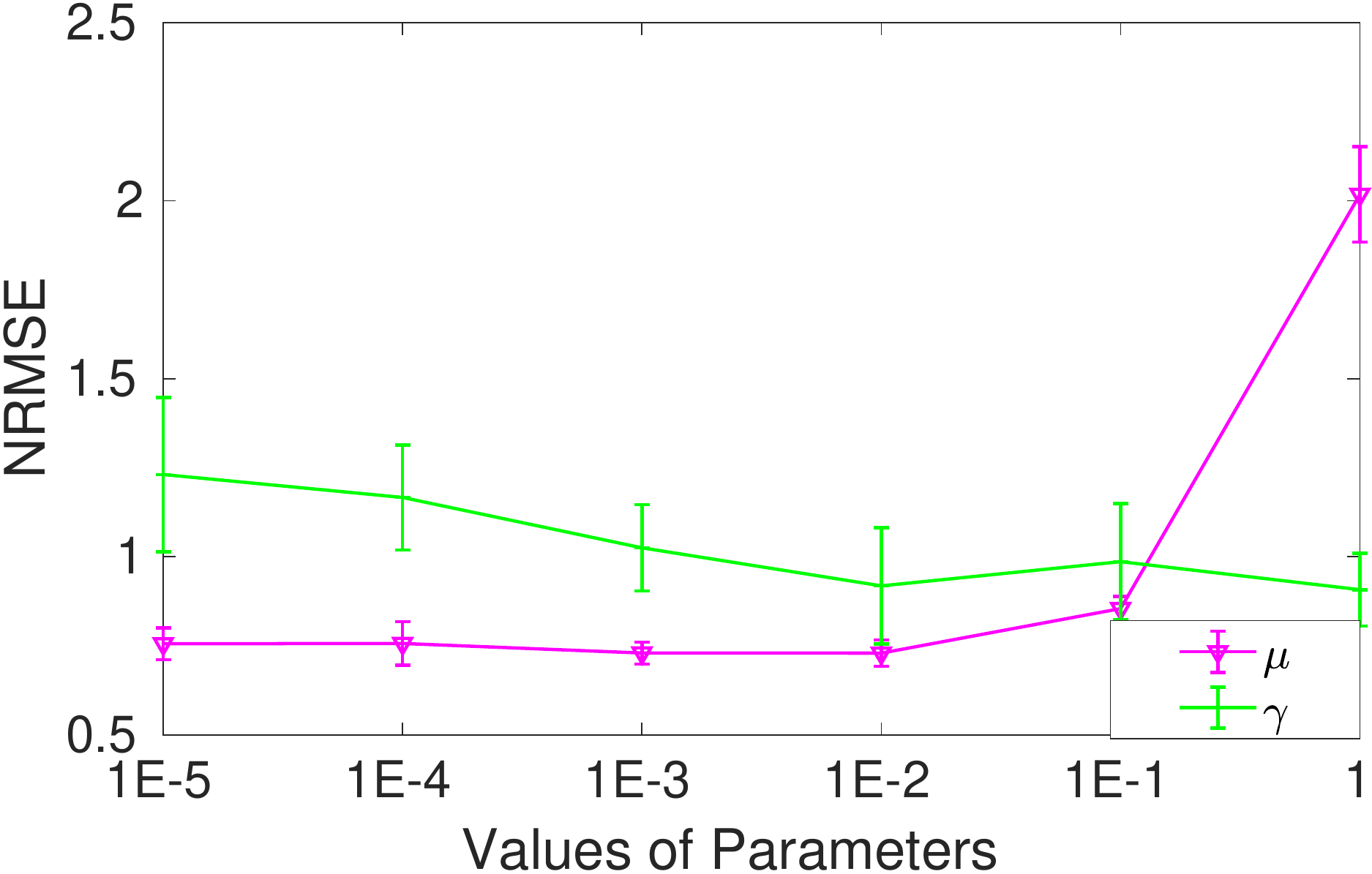}
  }
  \subfigure[\normalsize Model parameters on CH]{
  \includegraphics[width=0.47\columnwidth]{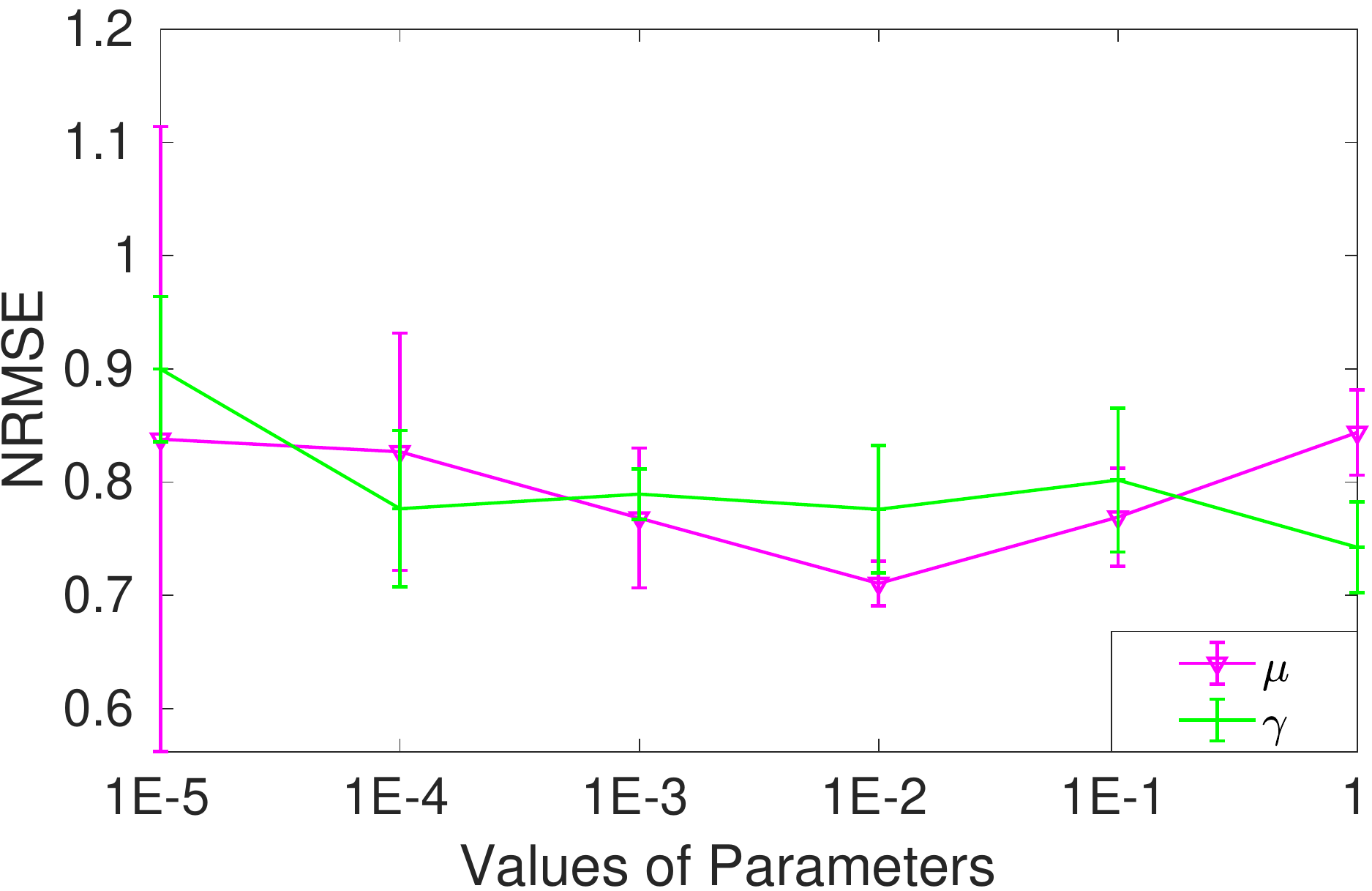}
  }
  \subfigure[\normalsize Model parameters on WMSL]{
  \includegraphics[width=0.47\columnwidth]{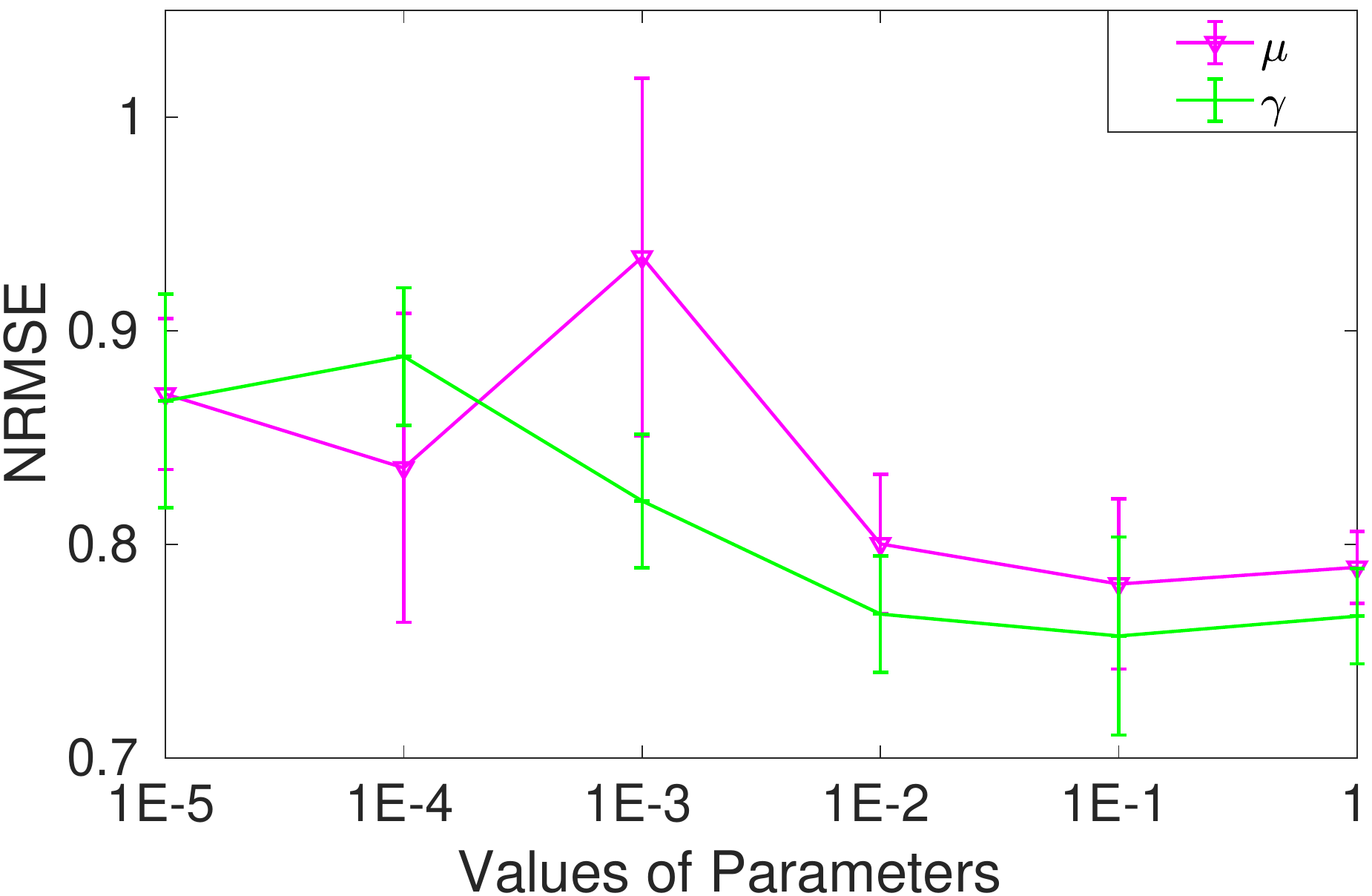}
  }
  \subfigure[\normalsize Model parameters on KCH]{
  \includegraphics[width=0.47\columnwidth]{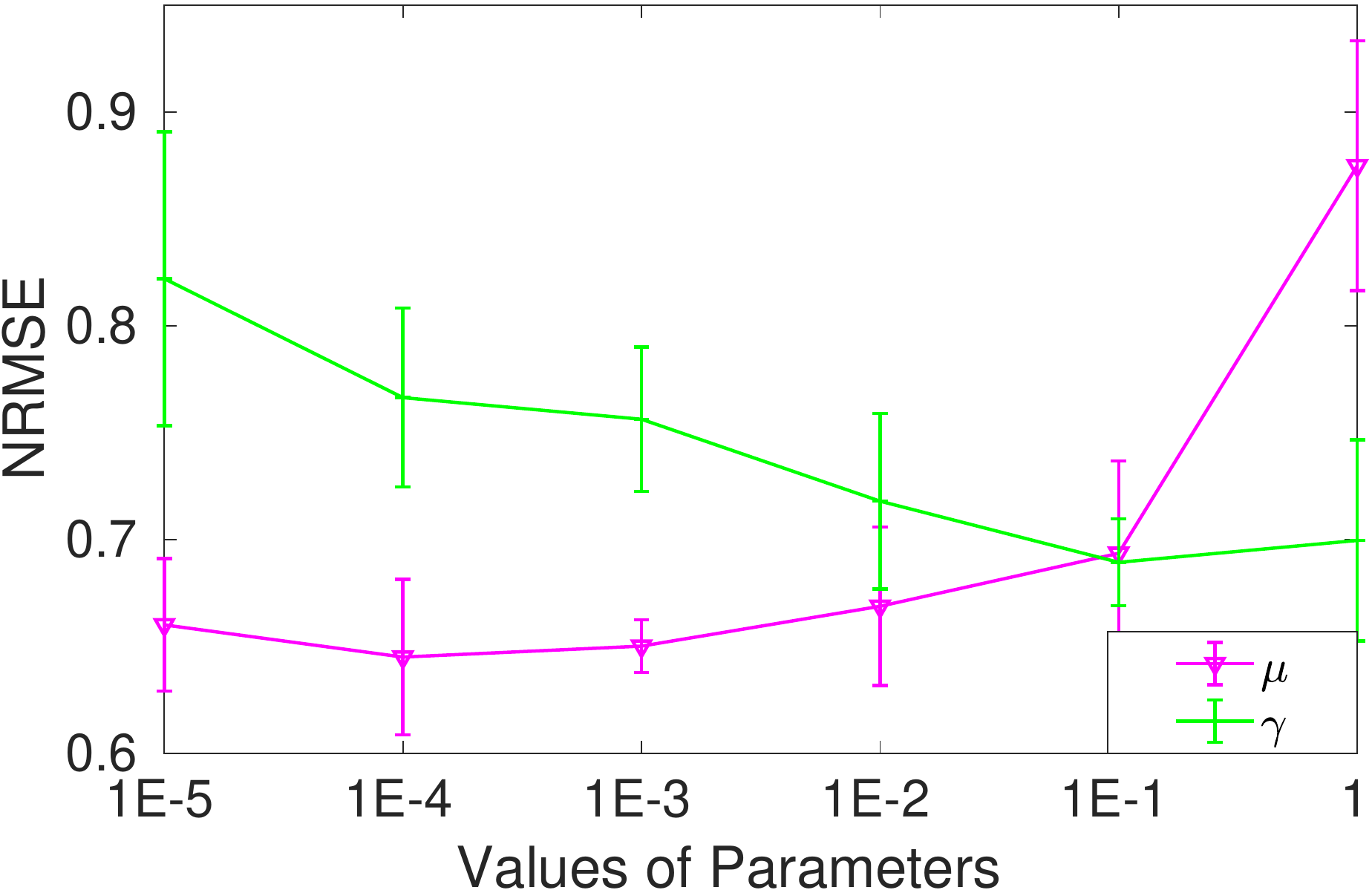}
  }
  \subfigure[\normalsize Laplacian parameters on Airfoil]{
  \includegraphics[width=0.47\columnwidth]{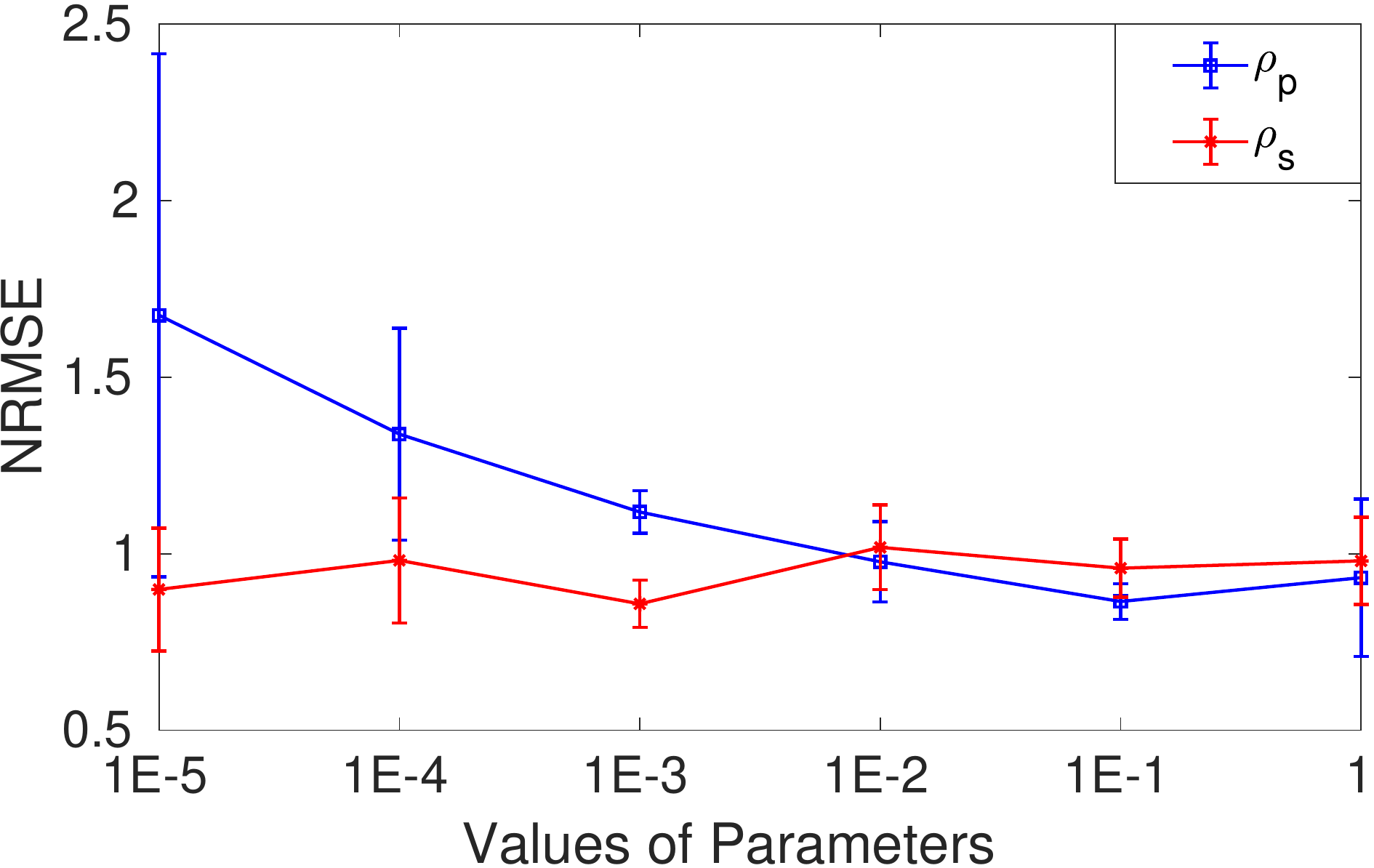}
  }
  \subfigure[\normalsize Laplacian parameters on CH]{
  \includegraphics[width=0.47\columnwidth]{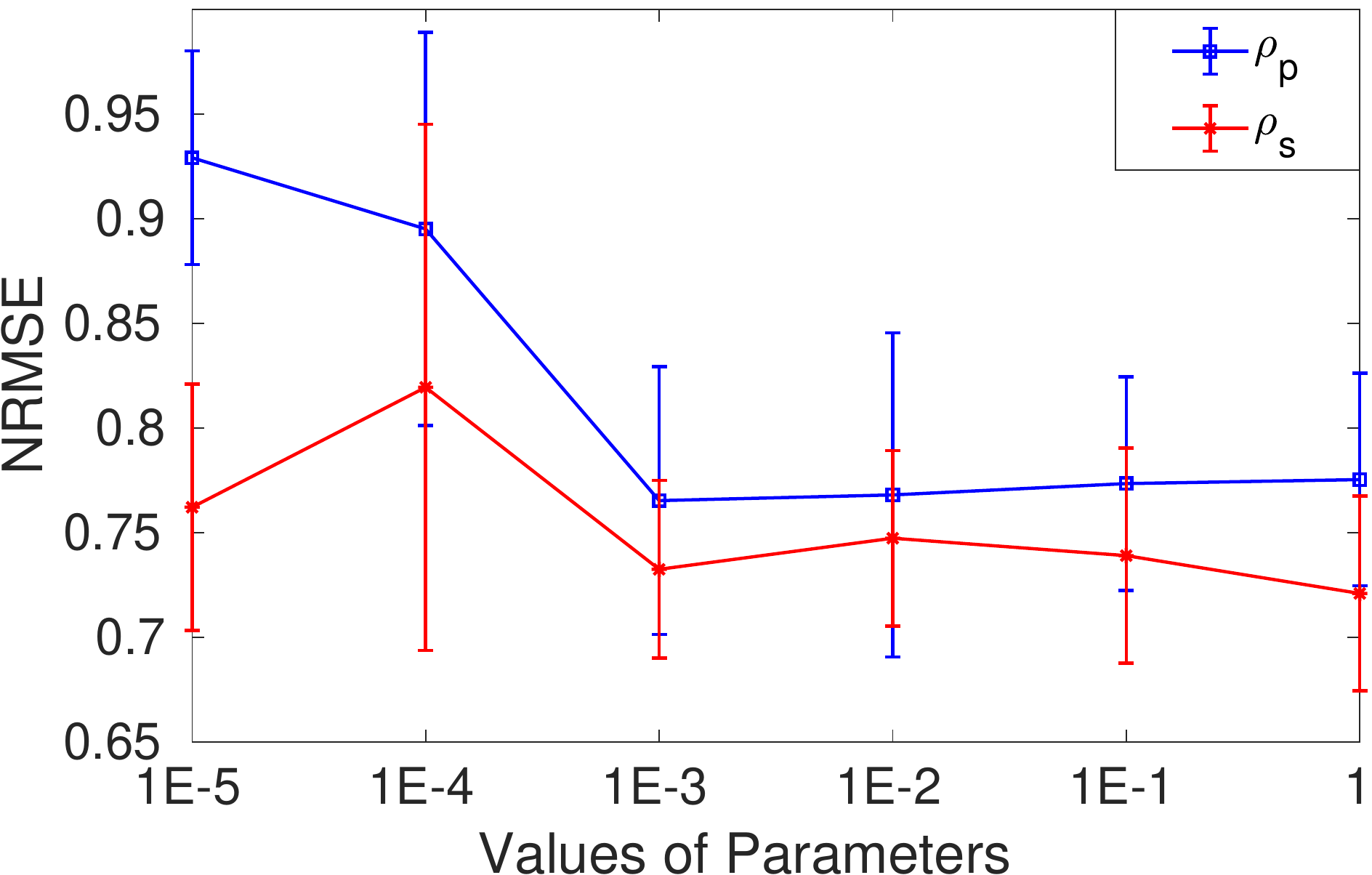}
  }
  \subfigure[\normalsize Laplacian parameters on WMSL]{
  \includegraphics[width=0.47\columnwidth]{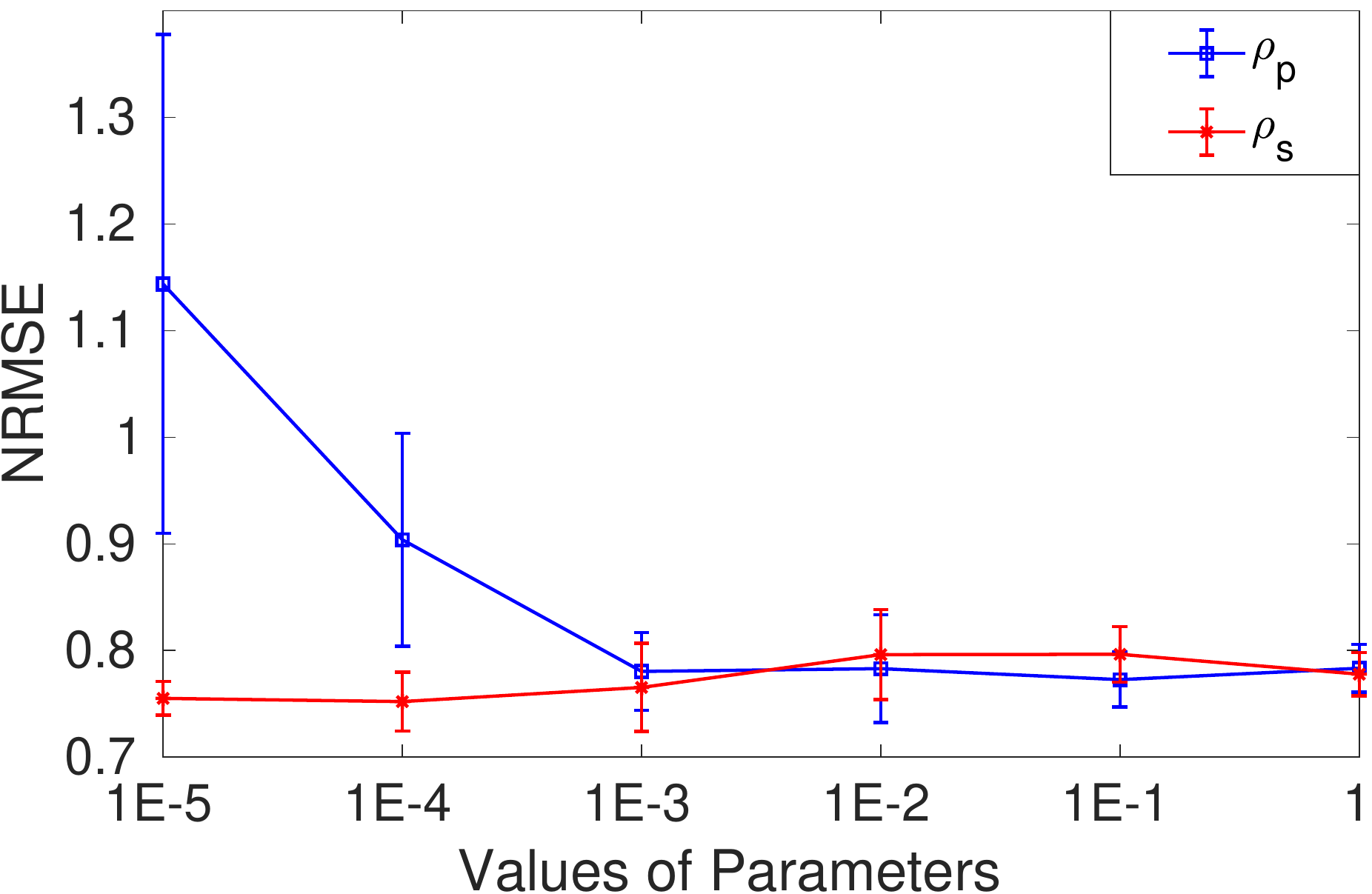}
  }
  \subfigure[\normalsize Laplacian parameters on KCH]{
  \includegraphics[width=0.47\columnwidth]{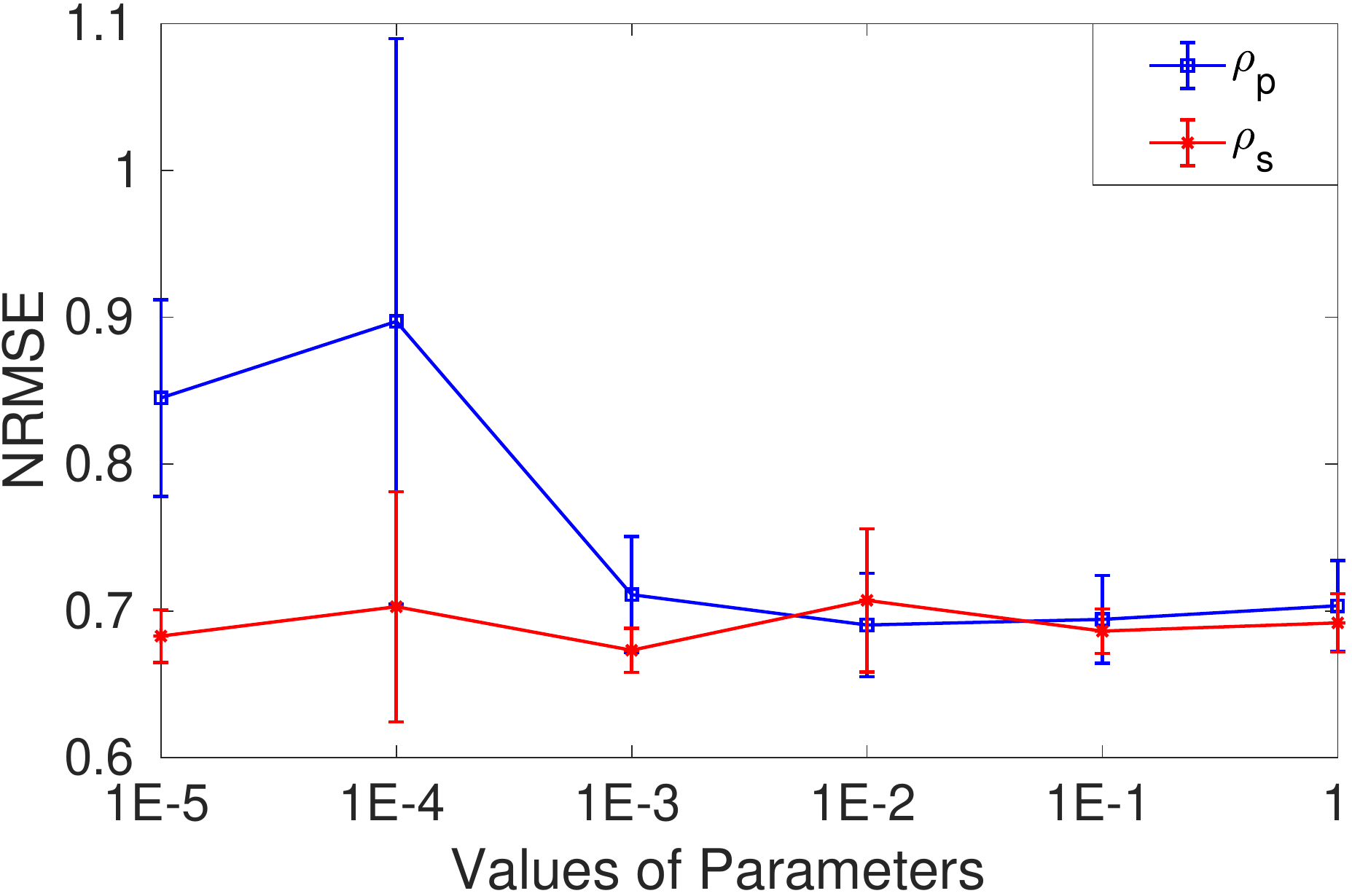}
  }
  \caption{\normalsize NRMSE and its standard deviation with different model parameters (top) and Laplacian parameters (bottom).}
  \label{fig:para_ana}
\end{figure*}

\begin{figure}[!ht]
  \centering
  \subfigure[\normalsize airfoil]{
  \includegraphics[width=0.45\columnwidth]{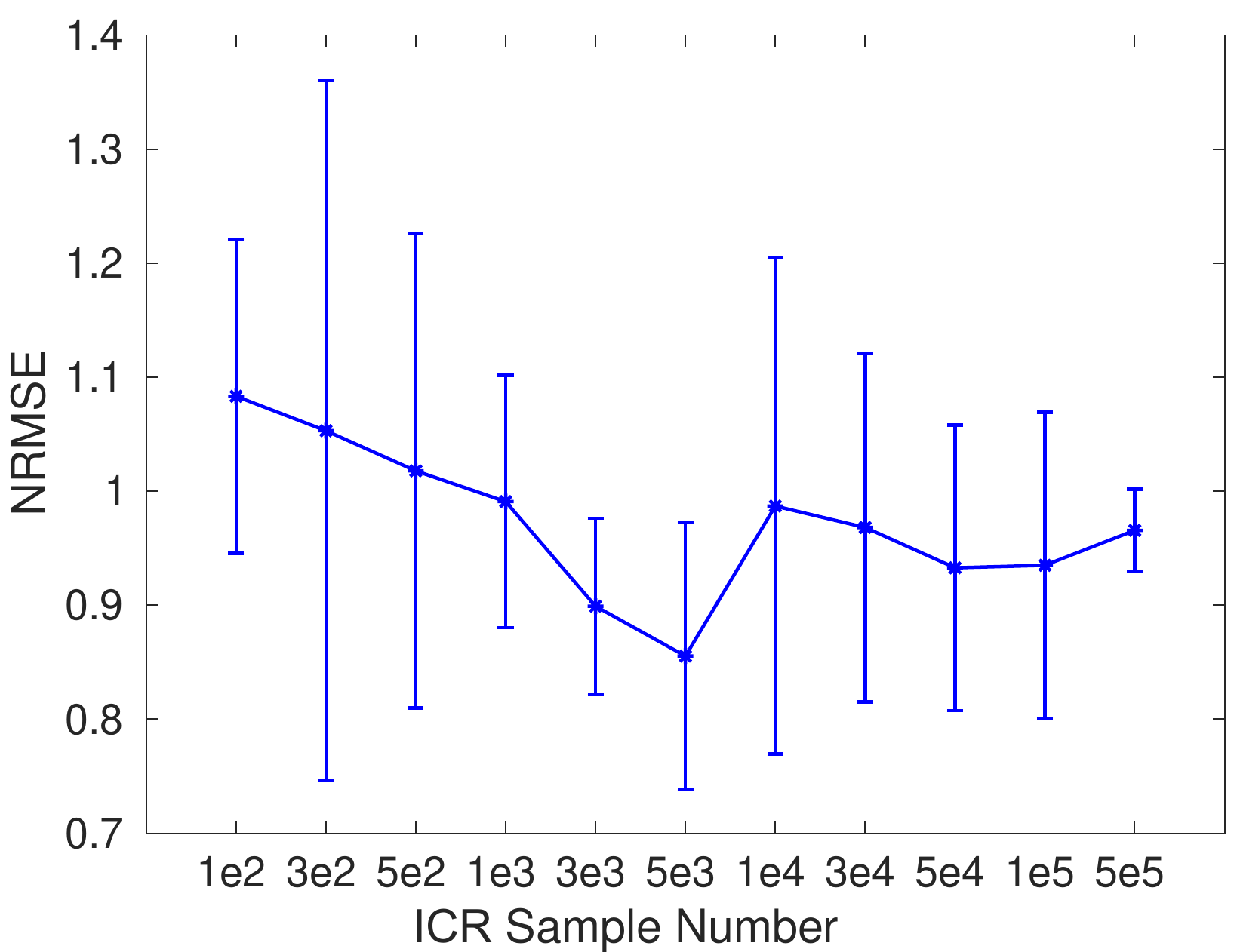}
  }
  \subfigure[\normalsize CH]{
  \includegraphics[width=0.45\columnwidth]{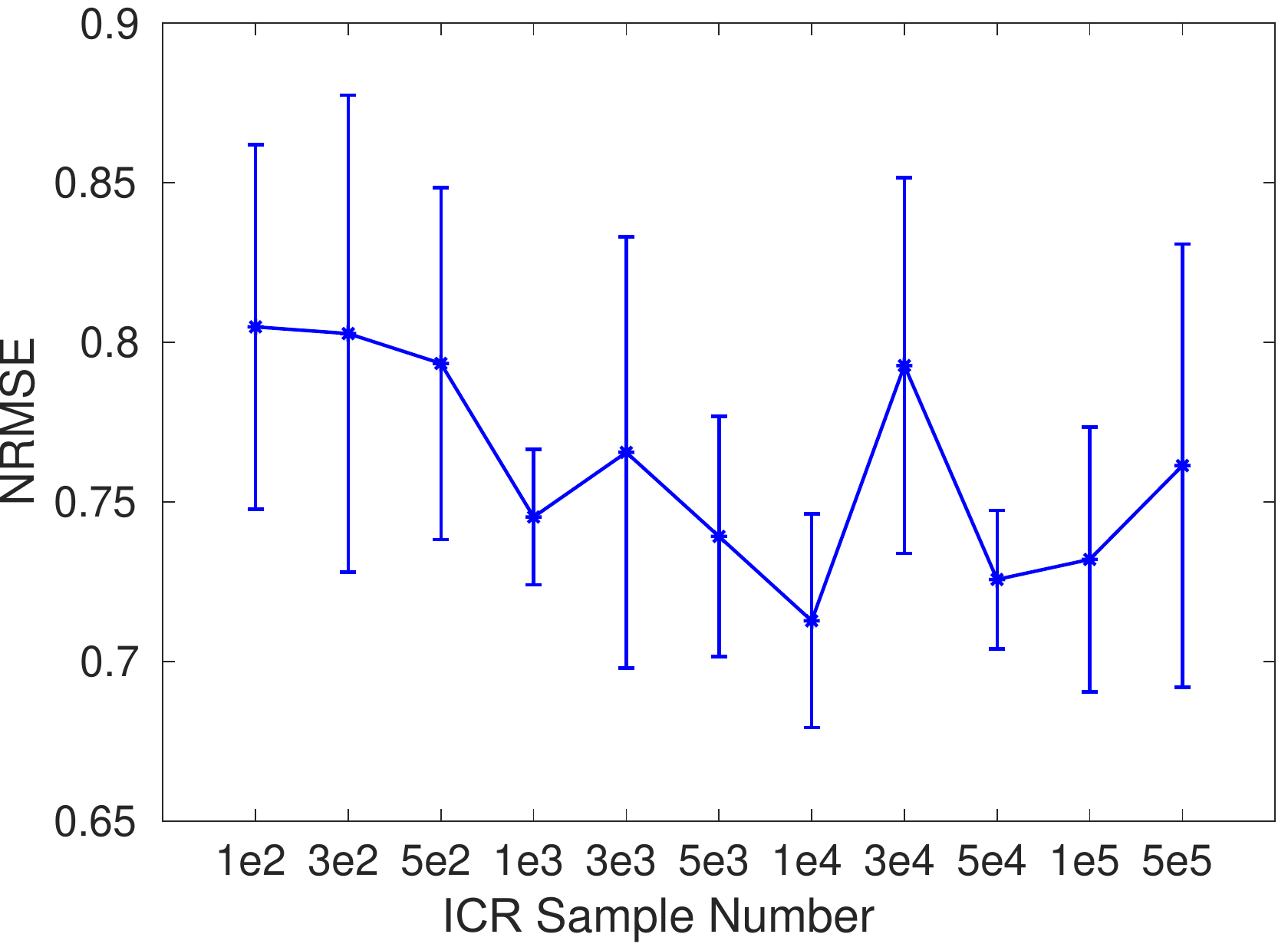}
  }
  \subfigure[\normalsize WMSL]{
  \includegraphics[width=0.45\columnwidth]{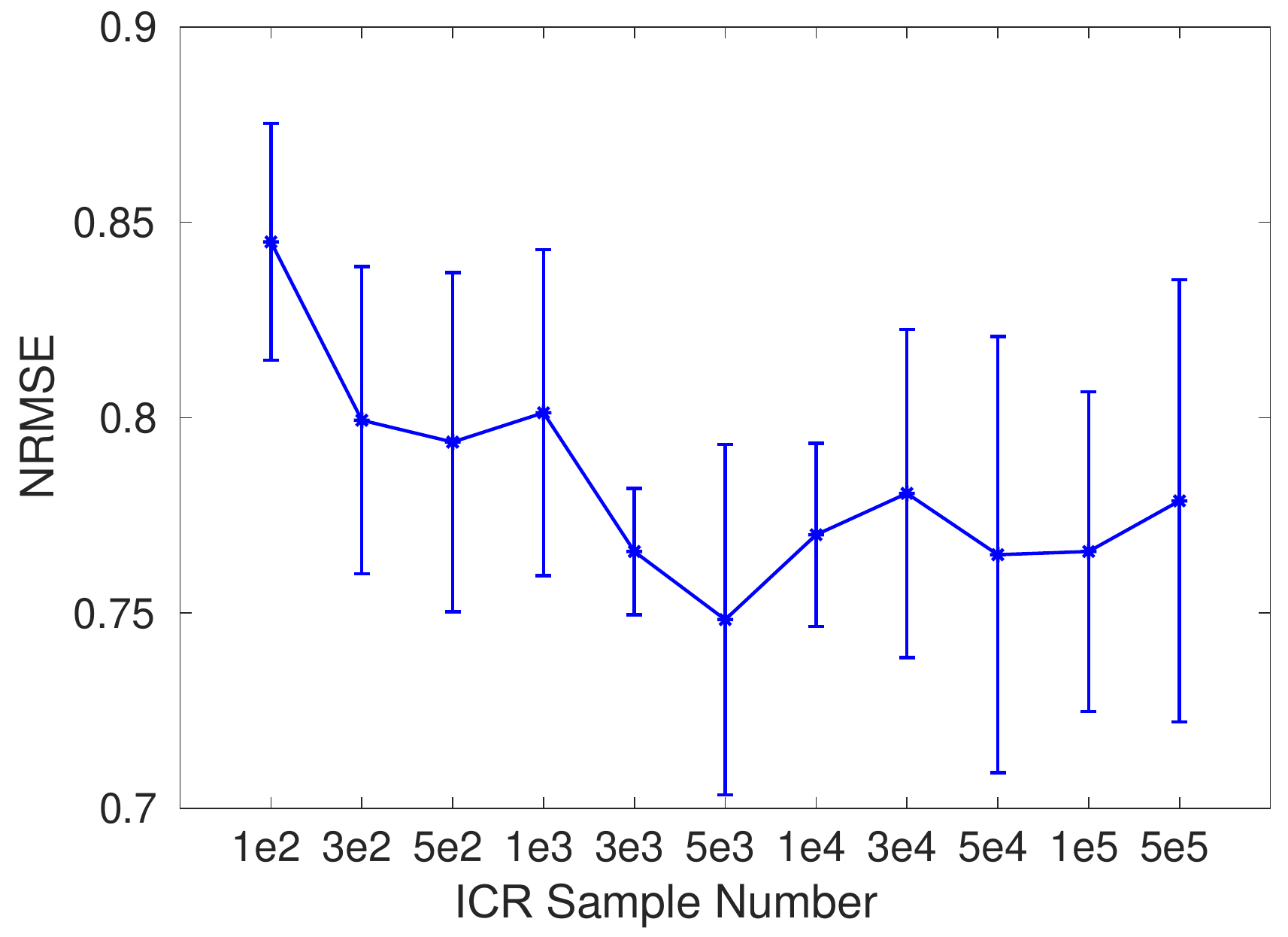}
  }
  \subfigure[\normalsize KCH]{
  \includegraphics[width=0.45\columnwidth]{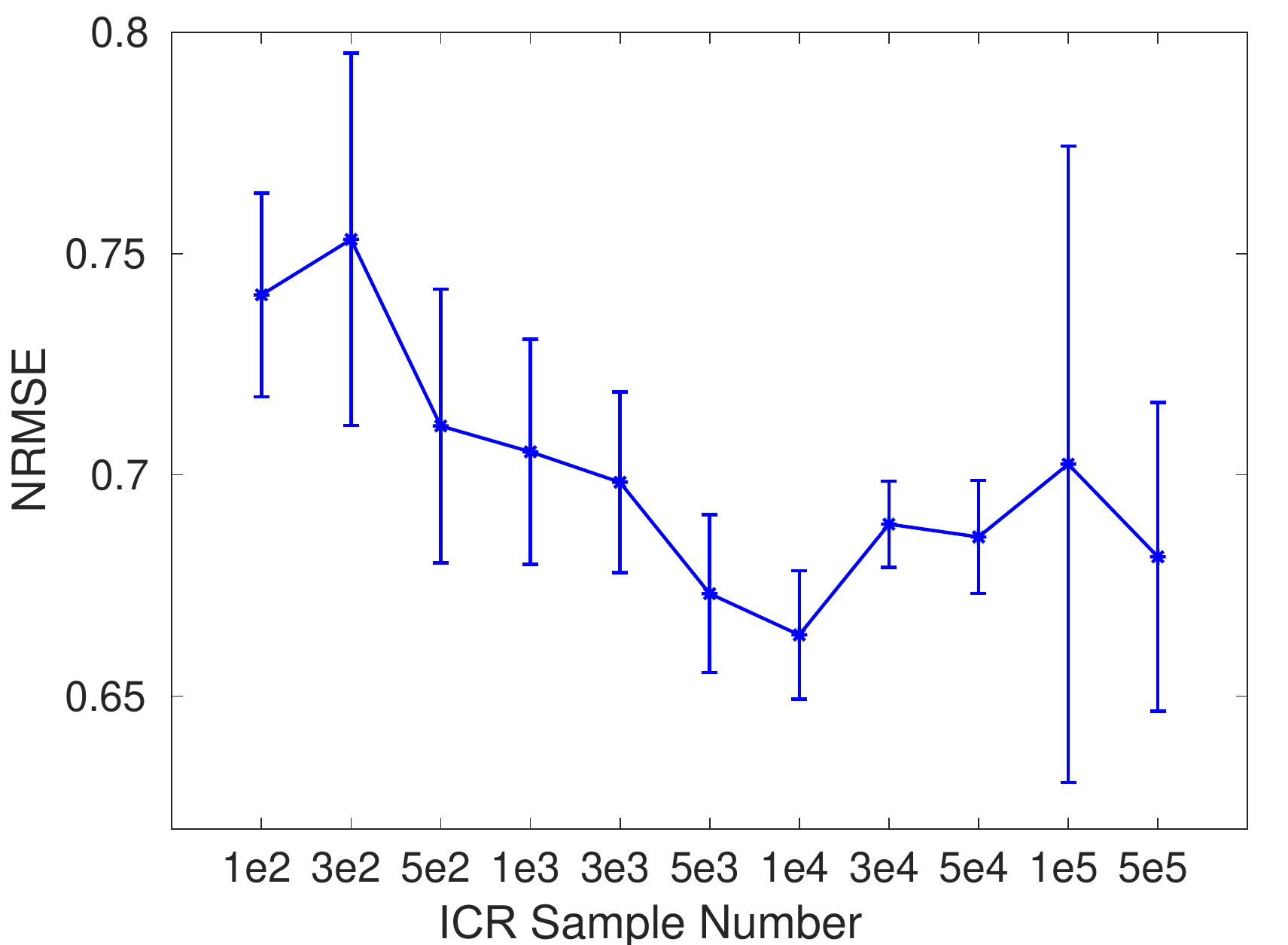}
  }
  \vspace{-10pt}
  \centering
  \caption{\normalsize NRSME and its standard deviation when varying the number of ICR samples.}
  \label{fig:mixup_ana}
\end{figure}

\begin{figure}[!ht]
  \centering
  \subfigure[\normalsize CH]{
  \includegraphics[width=0.45\columnwidth]{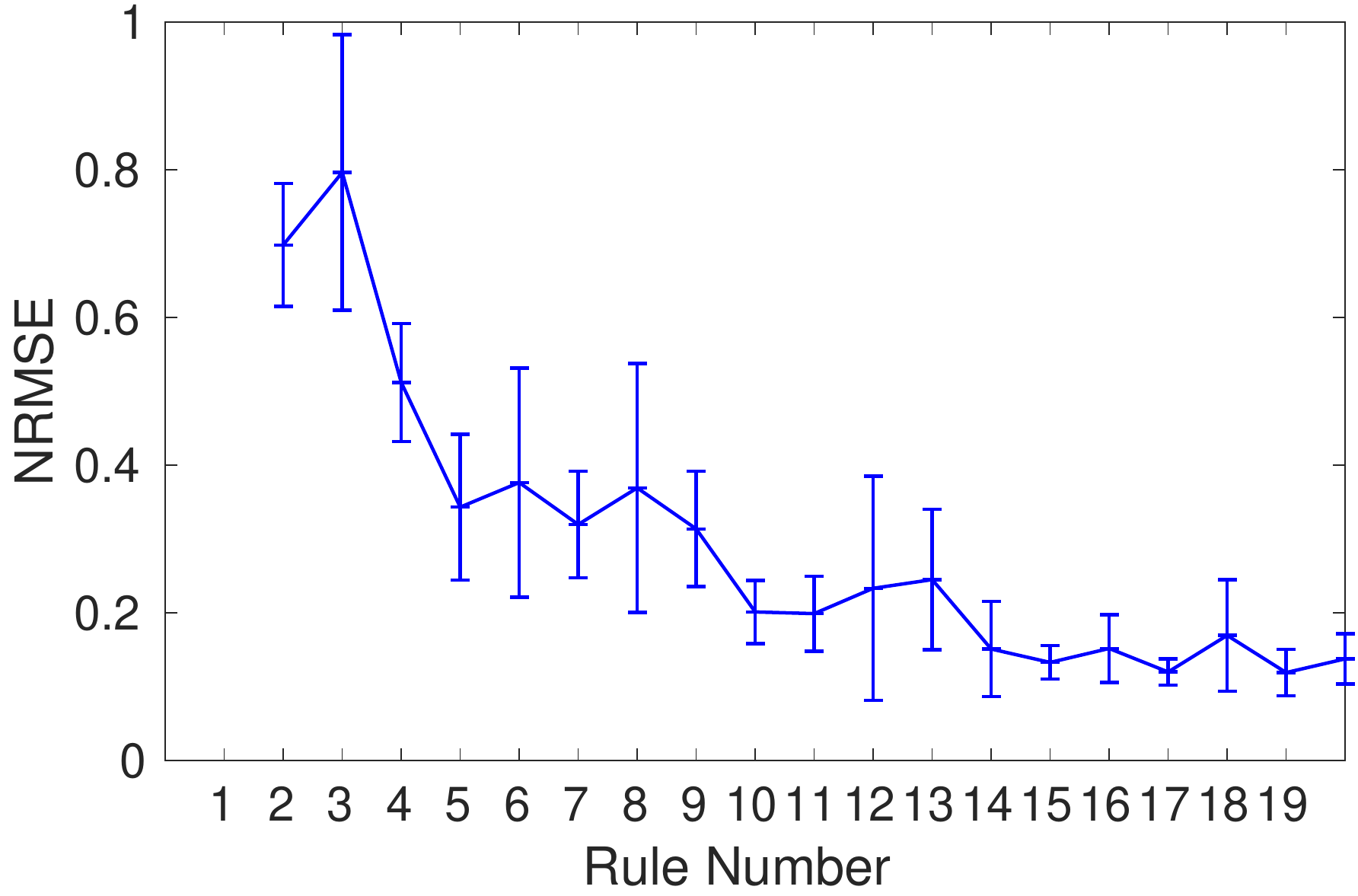}
  }
  \subfigure[\normalsize KCH]{
  \includegraphics[width=0.45\columnwidth]{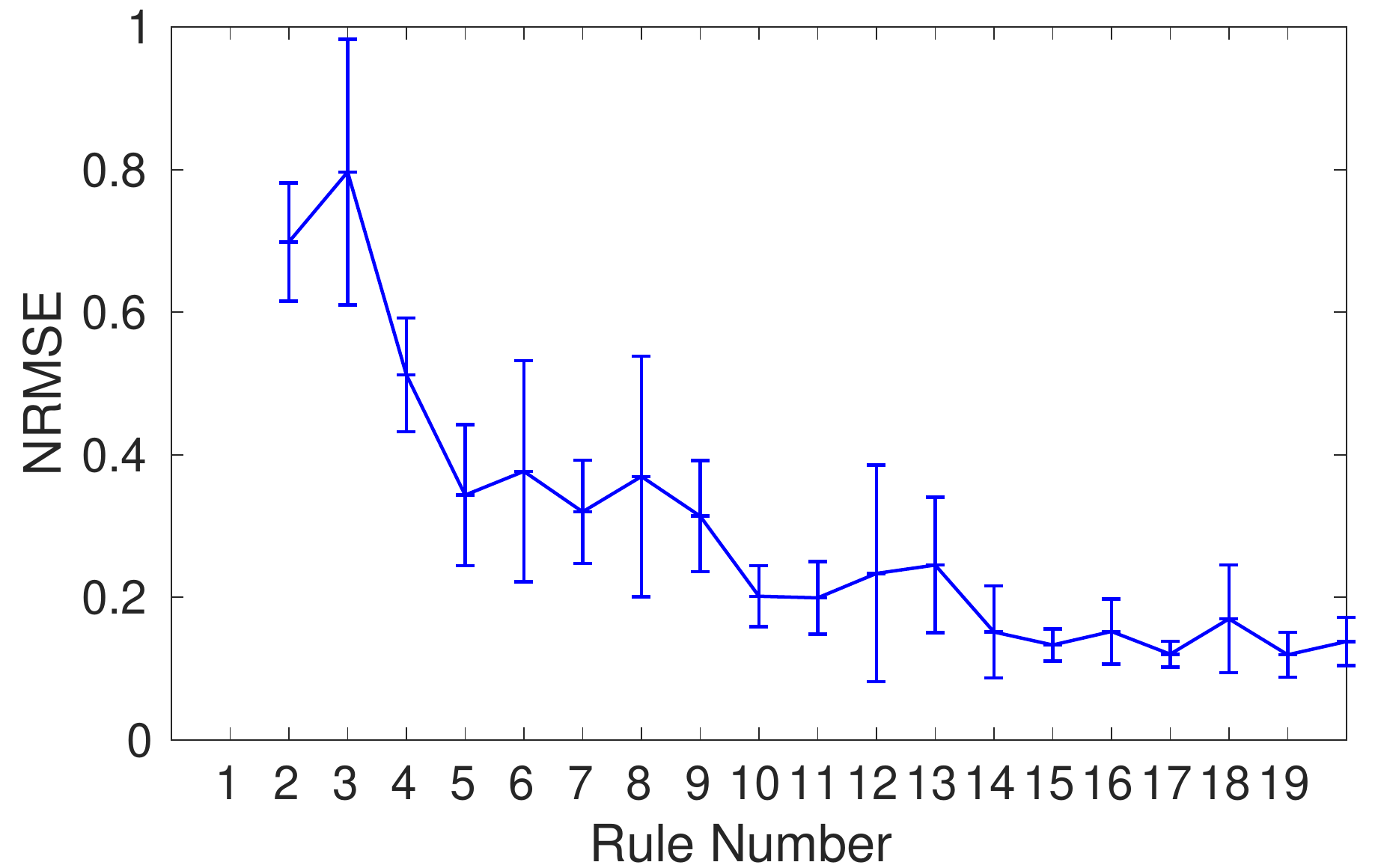}
  }
  \centering
  \caption{\normalsize NRSME and its standard deviation when varying the number of rules.}
  \label{fig:rule_ana}
\end{figure}

\begin{figure}[!ht]
  \centering
  \subfigure[\normalsize airfoil]{
  \includegraphics[width=0.45\columnwidth]{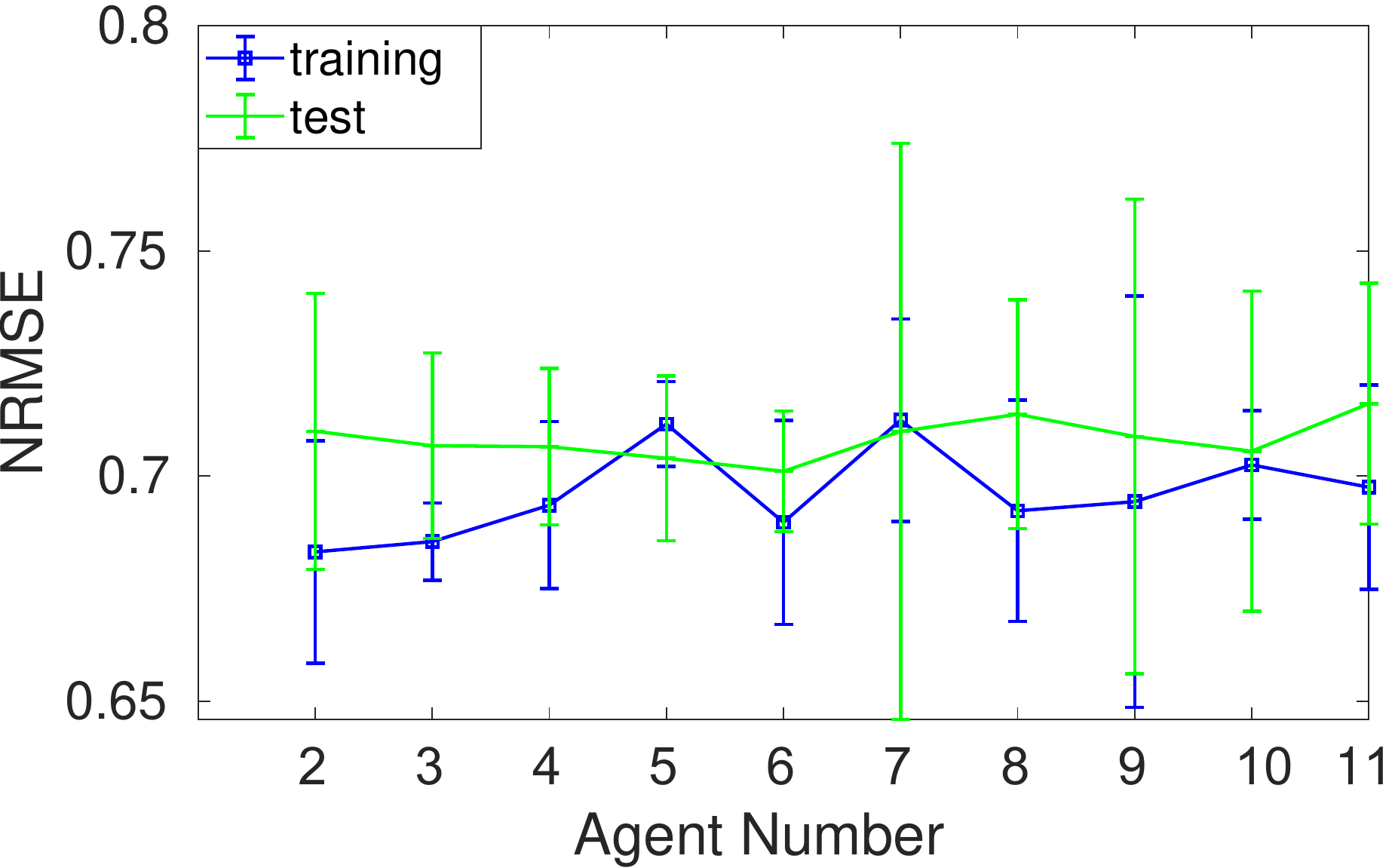}
  }
  \subfigure[\normalsize CH]{
  \includegraphics[width=0.45\columnwidth]{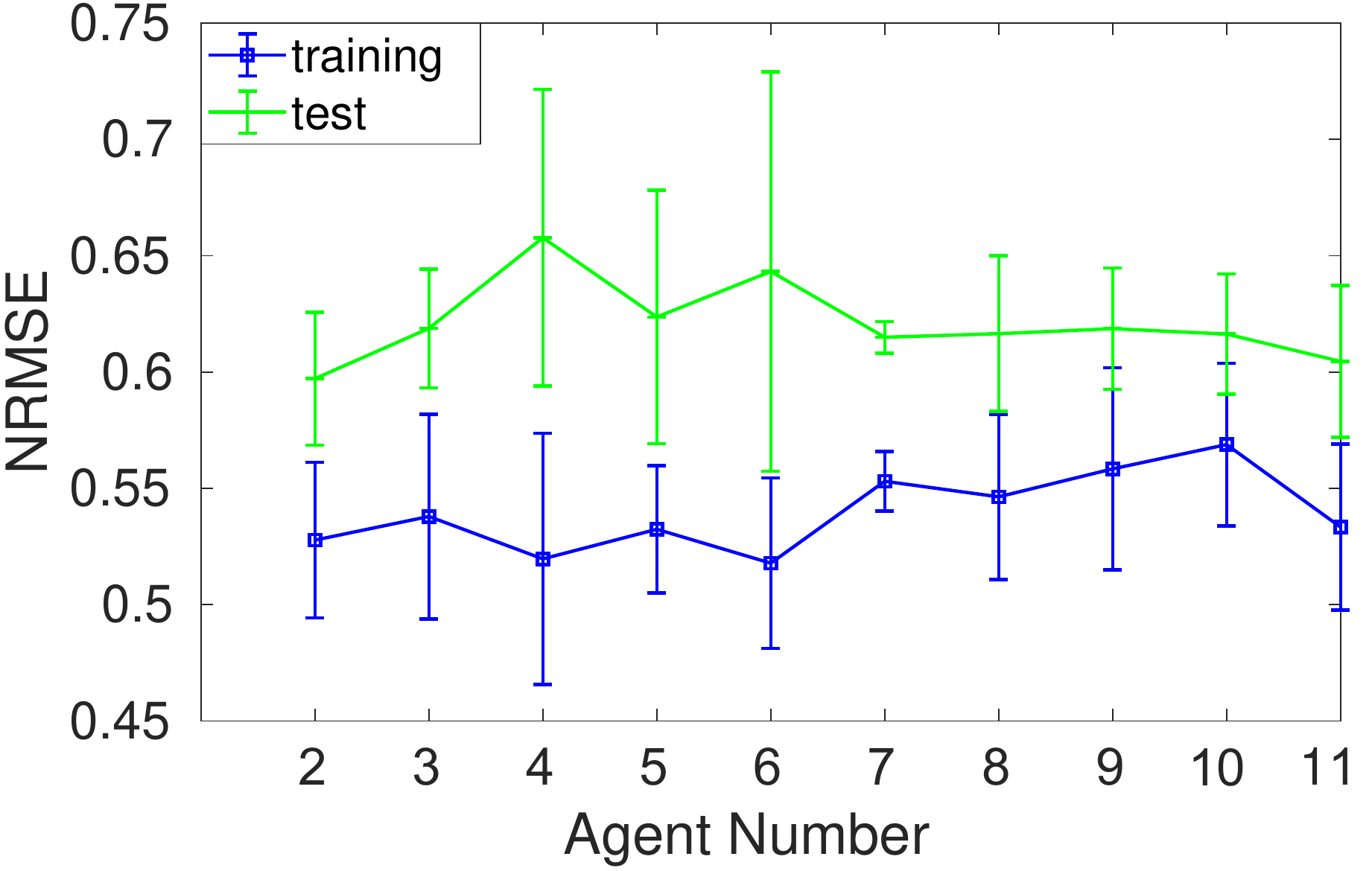}
  }
  \subfigure[\normalsize WMSL]{
  \includegraphics[width=0.45\columnwidth]{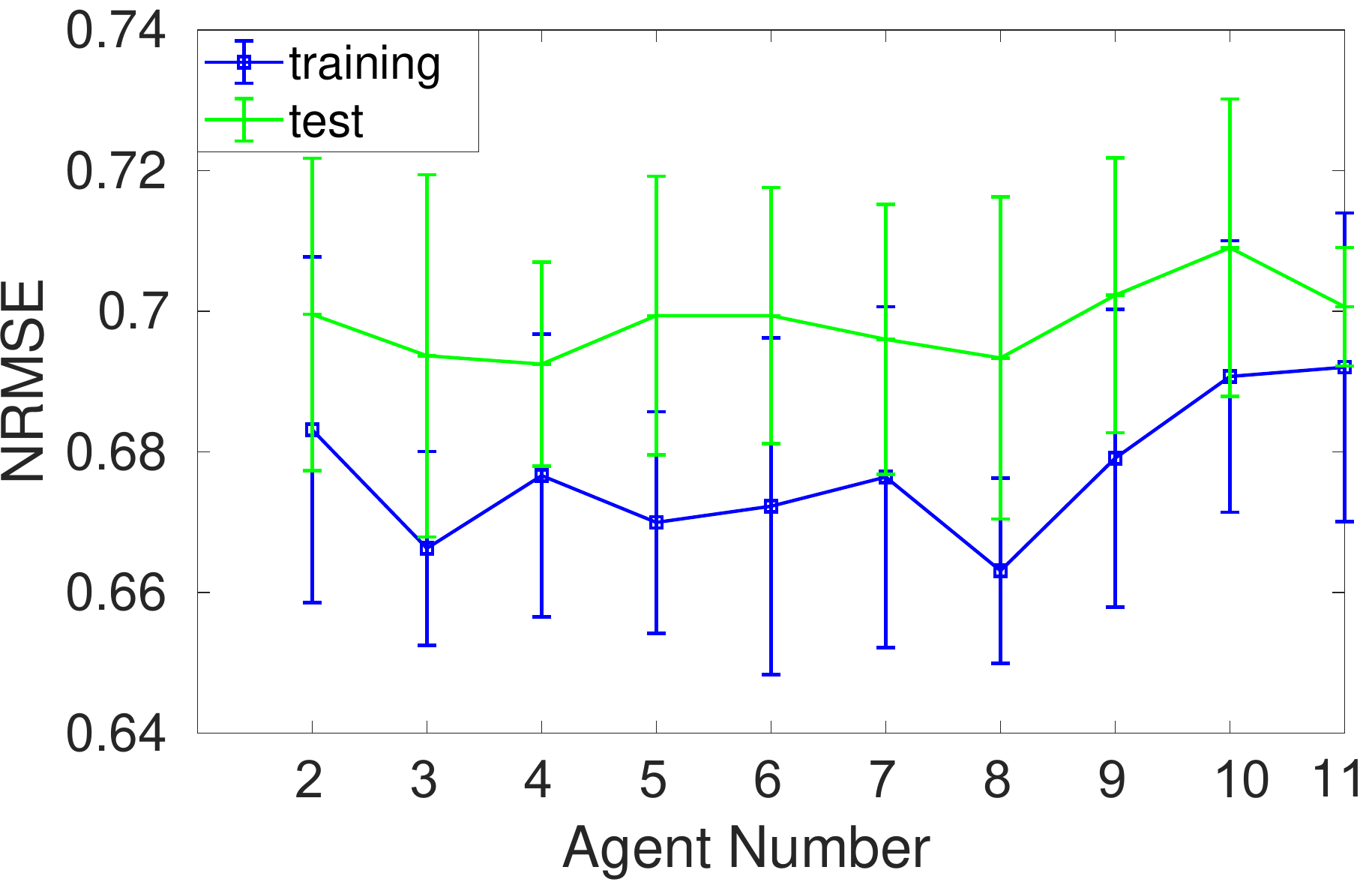}
  }
  \subfigure[\normalsize KCH]{
  \includegraphics[width=0.45\columnwidth]{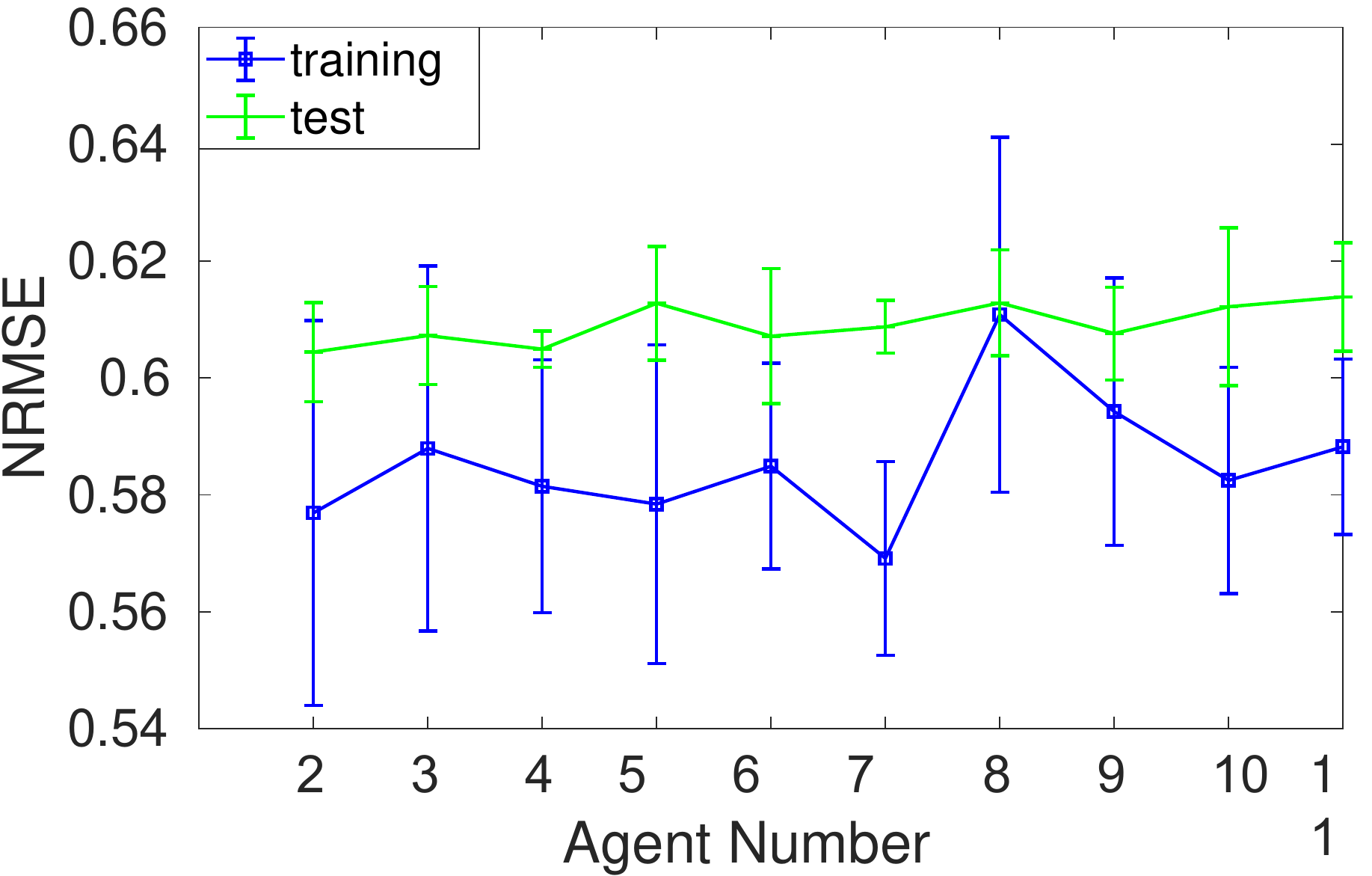}
  }
  \vspace{-10pt}
  \centering
  \caption{\normalsize NRSME and its standard deviation when varying the number of agents}
  \label{fig:agent_ana}
\end{figure}

\subsection{Performance on Different Datasets}
All parameters in this section were chosen from an interval of $\{10^{-5},10^{-4},10^{-3},10^{-2},10^{-1},1\}$. Further, the FCM parameter $\alpha$ was set to 1.1 for all experiments to increase the diversity of fuzzy rules. We repeated each experiment 10 times, reporting the average NRMSE and its standard variance as the results. For a fair comparison, we arranged the training sets on interconnected networks with five fully connected agents, setting the number of rules to 5 for all agents with all datasets. We used 50 samples in the centralized scenario and 10 samples for each agent in the distributed setting.

We compared several alternative methods within the semi-supervised fuzzy regression (SFR) framework that using different regularization, fully-supervised regression (FR) model, and one SSL model. All methods are compared in both centralized and distributed ways and described follows:
\begin{itemize}
\item[-] s-SFR: A variant of SFR, this method apply FCM method in the structure learning process to leverage both labeled and unlabeled samples. A $\ell_2$ regularization is used in the parameter learning process to enhance model generalization performance.
\item[-] \textbf{SFR-ICR}: Our proposed SFR model, based on s-SFR, this fuzzy regression model add an ICR term in the parameter learning process to further utilize unlabeled data.
\item[-] SFR-G: Another variant of SFR, different from SFR-ICR, this model involves a graph-based regularization term. We designed this method to show the efficiency and effectiveness of SFR-ICR under the same settings.
\item[-] FR: This is a fully-supervised fuzzy regression model. FR can provide a lower bound of the loss for the SFR-ICR.
\item[-] LapWNN: This method involves a graph-based SSL based on a wavelet neural network (WNN) \cite{xie2019distributed}. Since LapWNN was developed for distributed semi-supervised scenarios, thus it is a quite relevant baseline method.
\end{itemize}

The results shown in Table \ref{tbl:performance}, place SFR-ICR as the clear outstanding performer in both the centralized and distributed settings, providing a strong endorsement of the effectiveness of semi-supervised fuzzy regression. T-tests were implemented to show statistical significance between SFR-ICR and other methods. As shown in Table \ref{tbl:performance}, all the p-values are smaller than 0.05, thus the difference between our SFR-ICR and other methods are significant.

Notably, SFR-G outperformed LapWNN on all datasets in the distributed scenarios, indicating that FNN-based SSL outperforms WNN-based SSL methods. Particularly, SFR-ICR reduced 31.96\% NRMSE values compared with the distributed results by LapWNN on average of all datasets. Within the SFR scheme, the NRMSE is also smaller by SFR-ICR than by s-SFR and SFR-G, closely outpacing FR, which indicates that ICR has a much greater ability to extract useful information from unlabeled samples than other regularization methods. Particularly, SFR-ICR respectively reduced 23.19\% and 19.44\% NRMSE values compared with the distributed results by s-SFR and SFR-G on average of all datasets.

The SFR-ICR is also superior in efficiency. SFR-ICR costs far less time than other methods in centralized ways. For distributed methods, the reason computing speeds of some small-scale dataset is slower is that communication overhead among agents offsets the computing superiority of SFR-ICR.

We also assessed distributed SFR-ICR (DSFR-ICR) against its centralized counterpart SFR-ICR (CSFR-ICR). DSFR-ICR performed better on all datasets, except for California Housing, demonstrating that a distributed approach can further increase the model performance by spreading computing resources across different agents.

\subsection{Convergence Analysis}
To investigate the convergence behaviors of proposed algorithms, we plotted the loss values of DFCM and DICR with each iteration of the optimization procedure, as shown in Fig. \ref{fig:conv_ana}. In this and following subsections, we initialized the Laplacian parameter $\rho_p$ and $\rho_s$ at 0.1 and 0.1, and the parameters $\mu$, $\gamma$, and $\eta$, which is corresponding to the $\ell_2$  regularizer, the mix-up regularizer, and the graph-based regularizer, at 0.1, 0.1, and 0.0001, respectively. {\color{red} Due to space limitations, we randomly choose four datasets to show the results. DFCM converges within 25 iterations while DICR converges within 300 iterations for all the datasets. We can find that the required iteration number for DICR is quite larger than that for DFCM. The reason is that the training process of DICR is more complex than DFCM. Note that DICR needs to randomly generate virtual samples in a pair-wise manner to augment the training set. By this way, DICR can better exploit unlabeled samples but sacrifice the convergence speed.}

\subsection{Effects of regularization and ADMM parameters}
As is convention, we assessed the influence of each parameter by fixing all other parameters using the initialization mentioned in Section V.B and varying the parameter in appropriate intervals: $\{10^{-5},10^{-4},10^{-3},10^{-2},10^{-1},1\}$. The results of the parameter analysis is presented in Fig. \ref{fig:para_ana}.

In terms of the regularization parameters $\gamma$ and $\mu$, the NRMSE drops as parameter values increases, reaching its nadir at $\mu=1$ with all datasets. This suggests that ICR plays a critical role in the model.  The best result for $\mu$, however, differs for every dataset. On Airfoil, CH and KCH datasets, $\mu$ decreases before reaching its bottom, then starts to climb again, which indicates that the proportion of the $\ell_2$ term needs to be within a reasonable range.

The variance in the ADMM parameter $\rho_s$ at different values was very small. $\rho_p$ performs better with values greater than $10^{-3}$. Thus, we can conclude that $\rho_s$ is robust at a wide range of values, but $\rho_s$ performs better when choosing a relatively large value.

\subsection{Effects of the number of interpolated unlabel samples}
We also investigate how the number of interpolated
unlabel samples affect the performance of DICR. After fixing the other parameters, we set the interpolated sample number within the interval $\{100, 300, 500, 1K, 3K, 5K, 10K, 30K, 50K, 100K, 500K\}$. The results are plotted in Fig. \ref{fig:mixup_ana}. All curves show a trend that declines in the beginning then climbs after reaching a minimum NRMSE.  The best performance with the Airfoils and WMSL datasets was with 5K ICR MixUp samples. With CH and KCH, the best performance was at 10K samples. Hence, it is best to set the number of MixUp ICR samples near to the training sample numbers for each dataset.

\subsection{Effects of the rule and agent number}
The number of rules was varied in the interval of $\{2,3,\cdots,20\}$. As shown in Fig. \ref{fig:rule_ana}, the more rules, the better the performance. Here, we randomly selected 500 labeled training samples and distributed them relatively evenly over different numbers of agents. As shown in Fig. \ref{fig:agent_ana}, DSFR was able to stabilize with a range of agent numbers.

\section{Conclusion}
In this paper, we proposed a novel DSFR model with fuzzy if-then rules and ICR. The result is a framework that handles uncertainty and also dramatically reduces computation and communication overheads in interconnected networks with multiple agents. Adopting the ADMM strategy, DSFR involves a DFCM for structure learning and a DICR for parameter learning. Notably, both algorithms can extract useful information from not only labeled samples but also unlabeled ones. The DICR algorithm increases model performance and robustness by restricting the decision boundaries to low-density regions. Comprehensive experiments on both artificial and real-world datasets show that DSFR excels in efficiency and effectiveness compared to the state-of-the-art algorithms. A deep variant of DSFR will be considered in future work to handle high-dimensional unlabeled data. This method will modify fuzzy logics to make them more suitable to high-dimensional applications. In addition, we will extend the current distributed learning methods to suit for more complex distributed scenarios.

\ifCLASSOPTIONcaptionsoff
  \newpage
\fi

\bibliographystyle{ieeetr}
\bibliography{literatures}

\end{document}